\documentclass[sigconf, nonacm]{acmart}

\newcommand\vldbpagestyle{plain}

\usepackage[utf8]{inputenc} 
\usepackage[T1]{fontenc}    
\usepackage{hyperref}       
\usepackage{url}            
\usepackage{booktabs}       
\usepackage{amsfonts}       
\usepackage{nicefrac}       
\usepackage{microtype}      
\usepackage{xcolor}         
\usepackage{adjustbox}
\usepackage{algorithmic}
\usepackage[ruled,vlined,linesnumbered]{algorithm2e}
\usepackage{amsmath}
\usepackage{amsthm}
\usepackage{xspace}
\usepackage{enumitem}
\usepackage{thmtools} 
\usepackage{thm-restate}
\theoremstyle{plain}

\theoremstyle{definition}
\newtheorem{definition}{Definition}

\theoremstyle{remark}

\usepackage{cleveref}
\usepackage{wrapfig}
\usepackage{xspace}
\usepackage{enumitem}
\usepackage{mathtools}
\usepackage{multirow}
\usepackage{subfig} 

\usepackage{bm}

\DeclareMathOperator*{\argmin}{argmin}

\SetKwComment{Comment}{//\ }{}

\newcommand{\sys}{\texttt{TablePuppet}\xspace}

\begin{document}

\title{TablePuppet: A Generic Framework for Relational Federated Learning}

\author{Lijie Xu$^{\dagger}$, Chulin Xie$^{\ddagger}$, Yiran Guo$^{\mathsection}$,
Gustavo Alonso$^{\dagger}$, Bo Li$^{\natural}$, 
Guoliang Li$^{\mathparagraph}$, Wei Wang$^{\mathsection}$, \\ Wentao Wu$^{\flat}$, Ce Zhang$^{\natural}$}
\affiliation{
  \institution{$^{\dagger}$ETH Z\"urich \quad$^{\ddagger}$UIUC  \quad$^{\mathsection}$Institute of Software, Chinese Academy of Sciences \\ 
   \quad$^{\natural}$University of Chicago\quad$^{\mathparagraph}$Tsinghua University \quad$^{\flat}$Microsoft Research 
}
  \city{}
  \country{}
}

\begin{abstract}

Current federated learning (FL) approaches view decentralized training data as a single table, divided among participants either horizontally (by rows) or vertically (by columns). However, these approaches are inadequate for handling distributed relational tables  across databases. This scenario requires intricate SQL operations like \textit{joins} and \textit{unions} to obtain the training data, which is either costly or restricted by privacy concerns. This raises the question: \textit{can we directly run FL on distributed relational tables?}

In this paper, we formalize this problem as \textit{relational federated learning} (RFL). We propose \sys, a generic framework for RFL that decomposes the learning process into two steps: (1) \textit{learning over join} (LoJ) followed by (2) \textit{learning over union} (LoU). 
In a nutshell, 
LoJ pushes learning down onto the vertical tables being joined, and LoU further pushes learning down onto the horizontal partitions of each vertical table. 
\sys incorporates computation/communication optimizations to deal with the duplicate tuples introduced by joins, as well as \textit{differential privacy} (DP) to protect against both feature and label leakages.
We demonstrate the efficiency of \sys in combination with two widely-used ML training algorithms, \emph{stochastic gradient descent} (SGD) and \emph{alternating direction method of multipliers} (ADMM), and compare their computation/communication complexity.
We evaluate the SGD/ADMM algorithms developed atop \sys by training diverse ML models.
Our experimental results show that \sys achieves model accuracy comparable to the centralized baselines running directly atop the SQL results. 
Moreover, ADMM takes less communication time than SGD to converge to similar model accuracy.
\end{abstract}

\maketitle

\pagestyle{\vldbpagestyle}

\vspace{-0.5em}
\section{Introduction}

Federated learning (FL) is commonly used to train machine learning (ML) models on decentralized data~\cite{FL-DB, FL-Medical-DB, FEAST-SIGMOD, FL-DB-3, Yang-19-survey, VFL-survey, VFL-Review}. FL is particularly effective for scenarios where data sharing among participants is infeasible due to data privacy and security regulations like GDPR~\cite{GDPR}, or when data sharing is expensive, such as for geo-distributed or shared-nothing data systems~\cite{Fu-VLDB-cache, Teradata}.

Existing FL approaches, such as \textit{horizontal FL} (HFL)~\cite{FedAvg, HFL-1, HFL-2} and \textit{vertical FL} (VFL)~\cite{VFL-survey,VFL-Review}, view decentralized training data as a single, large table divided among participants, either by rows (HFL) or by columns (VFL). However, these approaches fall short in more common scenarios where training data is stored across relational tables in distributed databases, necessitating SQL operations like joins or unions to compose the training data. Take healthcare as an example, a patient's information can span different organizations' databases, such as hospitals, pharmacies, and insurance companies. Here, a patient can have multiple records in various hospital departments or pharmacy branches, considered as separate clients within these organizations. Data analysts need to aggregate the scattered tables using SQL operations (e.g., joins and unions) before model training. Given that VFL relies on one-to-one data alignment without joins and HFL is limited to tables with identical schemas, they are insufficient for this relational scenario. This leads us to ask: \emph{can we directly perform FL on these distributed relational tables without data sharing?}

In this paper, we call this problem \emph{relational federated learning} (RFL) and formalize it as FL atop \emph{union of conjunctive queries} (UCQ), a well-known notion in the database literature~\cite{Ullman88}. 
Relational tables in different organizations are referred to as \textit{vertical tables}, since they own different feature columns, albeit varying in size. 
Within the same organization, clients can have horizontal partitions of the vertical table, which we term as \emph{horizontal tables}.
To obtain the complete dataset for model training, one needs to union the horizontal tables and then join the vertical tables, which is \emph{conceptually} equivalent to performing a UCQ over all the tables involved.

RFL introduces several unique challenges. First, HFL/VFL approaches are inadequate for RFL, necessitating new learning methods. Research in both VFL and HFL~\cite{VFL-survey, FL-DB} highlights integrating SQL queries with FL, such as supporting one-to-many data alignment and federated databases, as an important future direction. Second,  given that RFL needs to be done conceptually on top of the UCQ result, which involves joining a large number of tables with duplicate tuples generated, the computation and communication overhead can be significant.  Third, unlike in VFL where data labels are owned by a single client, in RFL labels are horizontally partitioned and distributed across multiple clients, which requires mechanisms to ensure the privacy of both features and labels.

To address these challenges, we propose \sys, a generic framework for RFL that can be applied to two widely used ML training algorithms, \emph{stochastic gradient descent} (SGD) and \emph{alternating direction method of multipliers} (ADMM)~\cite{Boyd}.
\sys decomposes the RFL problem over the joined table (UCQ result) into two sub-problems: (1) \textit{learning over join} (LoJ) and (2) \textit{learning over union} (LoU). 
Essentially, LoJ pushes learning on the joined table down to the vertical tables being joined, and LoU further pushes learning on each vertical table down to each horizontal table of the union.
\sys optimizes and reduces the computation and communication overhead of duplicate tuples introduced by joins.
\sys also extends previous work on \emph{differential privacy} (DP)
for VFL~\cite{Chulin-ADMM} to establish privacy guarantees not only for the (distributed) features, but also for the (distributed) data labels.

The implementation of \sys adopts a server-client architecture, where the server can be one of the clients or an independent instance. 
The global ML model on the joined table is decoupled into local models held by clients. 
The server and clients collaboratively and iteratively train these local models. 
To unify the implementations of SGD/ADMM training processes, 
we design three physical operators in \sys that abstract their computation and communication: (1) a \emph{LoJ operator}, (2) a \emph{LoU operator}, and (3) a \emph{client operator} for the model updates inside clients. 
This abstraction facilitates the implementations of SGD/ADMM and the adjustment of computation/communication overheads in a flexible manner, leading to new SGD/ADMM algorithms that work not only for RFL (\textbf{RFL-SGD} and \textbf{RFL-ADMM}) but also for the vertical scenario of RFL without horizontal partitions (\textbf{RFL-SGD-V}, \textbf{RFL-ADMM-V}).

We study the effectiveness and efficiency of \sys, by evaluating model accuracy and performance of SGD/ADMM atop our \sys as well as counterparts in diverse scenarios. 
For model accuracy, we consider directly training centralized (i.e., non-federated) ML models on the joined table as the baseline approach. Our experiments show that \sys can achieve model accuracy comparable to this (strongest) baseline.
For performance, we focus on studying the communication overhead as it is the primary bottleneck in FL~\cite{FedAvg, HFL-2}, especially for geo-distributed clients. 
Experimental results show that ADMM atop \sys takes less communication time to converge (to similar accuracy) compared to SGD atop \sys. In particular, SGD/ADMM algorithms atop \sys outperform existing VFL methods that are forced to run directly on the vertical partitions of the
joined table, in terms of communication time.
The main contributions of this paper are as follows:

\begin{itemize}[leftmargin=2em]
    \item We propose a new RFL problem as  \emph{learning over the UCQ result} of relational tables across distributed databases.
    \item We present \sys, a generic RFL framework that pushes ML training down to individual tables, with both computation/communication optimization and privacy guarantees. 
    \item We implement \sys using a server-client architecture with an abstraction of three physical operators, which unifies the design and implementation of new SGD/ADMM algorithms for different scenarios of RFL.
    \item We provide a theoretical analysis of the computation and communication complexity of \sys, as well as its data privacy guarantees based on differential privacy.
    \item We evaluate \sys on real-world datasets with multiple ML models in diverse scenarios and our evaluation results demonstrate its effectiveness and efficiency.
\end{itemize}

\section{Relational Federated Learning}
\label{sec:problem-def}

\begin{figure}[t] 
\centering
\includegraphics[width=\columnwidth]{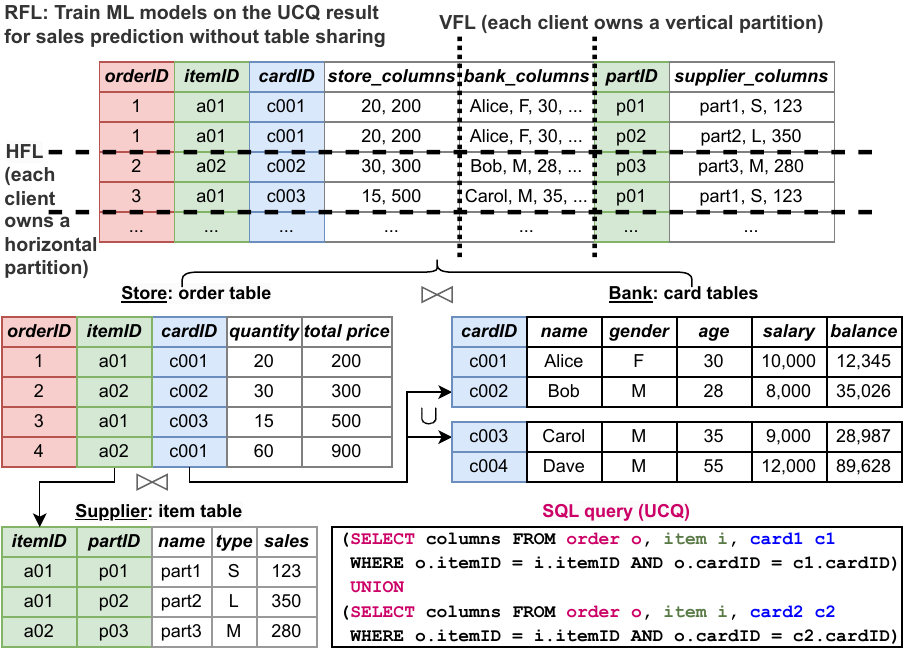}
\vspace{-1.5em}
\caption{A RFL example, using UCQ (with SQL syntax) to express the joins and unions over three tables.}
\label{UCQ-demo}
\end{figure}

\subsection{Example Scenario of RFL}

\begin{figure*}[th]
\centering 
\includegraphics[width=\textwidth]{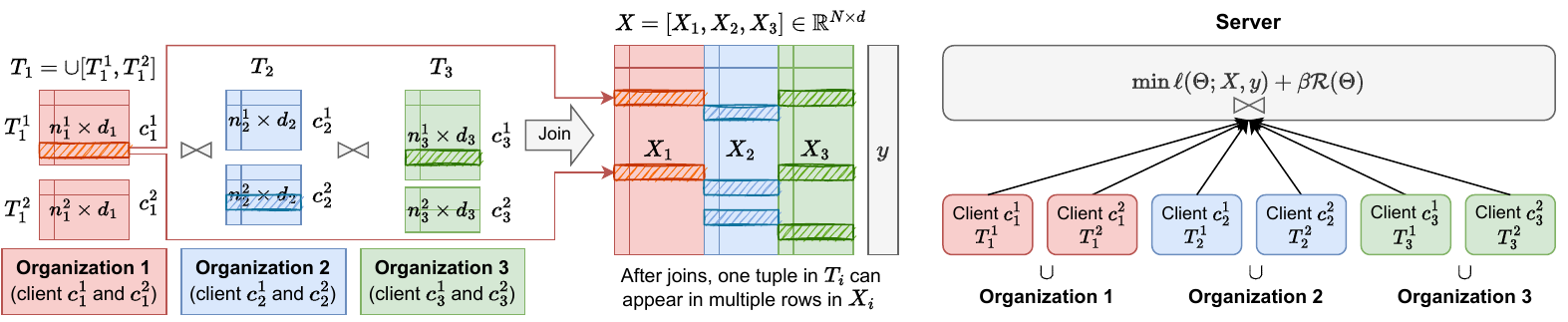}
\vspace{-1.5em}
\caption{Illustration of RFL where the table $X$ with column $y$ is the UCQ result. 
Each horizontal table $T_i^q$ is owned by a client $c_i^q$.}
\label{Fg:Join-VFL-Intro}
\end{figure*}

Assume there are three organizations, a store, a supplier, and a bank, as well as the following FL task: the store wants to use both the supplier's and the bank's data to train a user purchase model for sales prediction. 
As shown in Figure~\ref{UCQ-demo}, 
the store owns an \textbf{order} table and the supplier owns an \textbf{item} table,
where an item contains one or multiple parts. 
Moreover, the bank owns a \textbf{card} table.
To obtain the complete training data for the FL task, one needs to \emph{join} the three tables \textbf{order}, \textbf{item}, and \textbf{card} with the join predicate
$$\textbf{order}.\textit{itemID} = \textbf{item}.\textit{itemID}\quad\textbf{and}\quad\textbf{order}.\textit{cardID} = \textbf{card}.\textit{cardID}.$$ 
The join columns can contain duplicates.
For example, as shown in Figure~\ref{UCQ-demo}, the \textbf{order} table contain multiple tuples with the same \textit{itemID} or the same \textit{cardID}. 
Consequently, the join result can be substantially larger than each individual table.

Each table involved in the join can be \emph{horizontally} partitioned into multiple tables, and each horizontal table can be owned by a different client.
In the example shown in Figure~\ref{UCQ-demo}, the \textbf{card} table is partitioned into two tables, $\textbf{card}_1$ and $\textbf{card}_2$, that are managed by two branches of the same bank or even by two distinct banks maintaining the same schema as the original \textbf{card} table. 
This necessitates an additional \emph{union} over the join results.
In database literature, it can be naturally expressed as \emph{union of conjunctive queries} (UCQ)~\cite{Ullman88}.

\subsection{Problem Formalization} \label{sec:problem-def}

Assume there are $M$ organizations and the $i$-th organization consists of $Q_i$ clients. These $Q_i$ clients jointly possess a vertical table $T_i$, where the $q$-th client possesses a horizontal part of table $T_i$ denoted as $T_i^q$ and thus $T_i = \cup[T_i^1, \dots, T_i^{Q_i}], \forall i \in [M]$. We call $T_i^{q}$ a \textit{horizontal table} of $T_i$. Each $T_i^{q}$ contains $n_i^q$ tuples and $d_i$ features, and therefore
$T_i$ contains $n_i = \sum_{q=1}^Q{n_i^q}$ tuples and $d_i$ features (i.e., $T_i \in \mathbb{R}^{n_i \times d_i}$). The total client count is $Q$, which is the sum of $Q_i$ across all organizations. We can perform SQL (UCQ) queries on these tables $\{T_i^q\}_{q\in[Q_i], i \in [M]}$,
as shown in Figure~\ref{Fg:Join-VFL-Intro}. We assume that one of the $\{T_i\}_{i \in [M]}$ tables contains the \emph{label} column.

We denote the entire \emph{joined table} (i.e., the UCQ result) as $X = \,\bowtie[T_1, \dots, T_M] = [X_1, \dots, X_M]$, 
which contains $N$ tuples, $d=\sum_{i=1}^M d_i$ feature columns, and a label column $y \in \mathbb{R}^N$.
A single server is responsible for coordinating these clients to jointly train machine learning models on $X$.
We can now define the optimization problem associated with RFL as Eq.~\ref{Eq:HybridFL-problem}: 

\begin{gather} 
\min_{\Theta}\frac{1}{N}  \sum_{j=1}^N \ell(\Theta;X_{i,j}, y_j) + \beta \mathcal{R}(\Theta),  \label{Eq:HybridFL-problem}
\\ \text{where } X_{i,j} = \{\,\bowtie[T_1, \dots, T_M]\}_{i,j}, T_i = \cup[T_i^1, \dots, T_i^{Q_i}], i \in [M]. \nonumber
\end{gather}
Here, we assume that the server has obtained and stored the UCQ results as $X = [X_1, \dots, X_M] = \{X_{i,j}\}_{i \in [M], j \in [N]}$, i.e., $X_{i,j}$ is the $j$-th tuple of $X_i$.
In this case, we can directly train the ML models with parameters $\Theta$ on $X$ in the server. 
Moreover, $\ell$ represents the \emph{loss function} (e.g., cross entropy loss) and $\mathcal{R}$ represents the \emph{regularization function} (e.g., $L^2$-norm) for $\Theta$ (with constant $\beta$). In practice, clients do not share tables with the server, so the server cannot directly obtain $X$. As a result,
Eq.~\ref{Eq:HybridFL-problem} \emph{cannot be directly solved as it is.}

\subsection{Limitations of Existing Work} \label{sec:existing-FL}

We present a brief overview of existing work on horizontal and vertical FL, and discuss its limitations for the RFL problem.

\textbf{Horizontal FL (HFL)}
is a special case of RFL when there is only one organization (i.e., $M = 1$). Each client in the organization owns a horizontal partition of a large table~\cite{FedAvg, HFL-1, HFL-2}.
Existing work for HFL, such as FedAvg~\cite{FedAvg}, usually trains local models atop individual horizontal partitions and then synchronizes (e.g., by taking the average of) the model parameters or gradients periodically.

\textbf{Vertical FL (VFL)} \label{sec:vfl} is another special case of RFL. A single table is shared across multiple clients where each client owns a \emph{vertical partition}, i.e., a set of feature columns.
Moreover, it assumes that the tuples in each vertical partition can be one-to-one aligned without joins.
VFL has been extensively studied~\cite{VFL-survey, cheng2021secureboost, wu2020privacy, gu2020federated, hardy2017private,yang2019parallel,zhang2021secure, feng2020multi, hu2019learning, liu2019communication, vepakomma2018split,chen2020vafl,hu2019fdml,jin2021catastrophic}, and typically leverages SGD~\cite{VFL-survey, hu2019fdml, chen2020vafl} or ADMM~\cite{Chulin-ADMM, hu2019learning} to train local models in distributed clients.

\textbf{Relational FL (RFL)} is more complex than HFL and VFL, with both vertical and horizontal tables of different sizes, as well as one-to-many or many-to-many relationship introduced by joins. 
We cannot directly run HFL/VFL on the horizontal tables shown in Figure~\ref{Fg:Join-VFL-Intro}, because we cannot directly define the model objective functions for HFL/VFL on these horizontal tables (e.g., $T_1^1$ and $T_1^2$), as the model objective function is defined on the joined table.  
If we want to use HFL/VFL as shown in Figure~\ref{UCQ-demo}, we need to perform the UCQ to extract the training data and then repartition the UCQ result to fit into the data layout expected by the existing HFL/VFL methods. However, it requires table sharing that is infeasible.
In the rest of this paper, we present \sys, a novel generic framework for RFL that generalizes HFL and VFL with SQL operations.

\section{Overview of \sys} \label{sec:vertical-table-ADMM-framework}

The key idea of \sys is to \emph{decompose} the learning process involved in RFL into two steps: (1) \textit{learning over join} (LoJ) followed by (2) \textit{learning over union} (LoU).

Specifically, the LoJ step pushes ML on the entire joined table to each (virtual) vertical table $T_i$ instead of the vertical partition of the joined table. This step involves a \emph{table mapping mechanism} to avoid actual joins and unions, and it employs optimization strategies that can significantly reduce the computation and communication complexities (from $\mathcal{O}(N)$ to $\mathcal{O}(n_i)$, where $N$ is the length of the joined table and $n_i$ denotes the length of each vertical table $T_i$). For each learning problem on the vertical table $T_i$ resulted from the LoJ step, the LoU step further pushes ML computation to each horizontal table $T_i^q$.

\sys coordinates the LoJ and LoU computation on vertical/horizontal tables using a server-client architecture (detailed in Section~\ref{sec:ADMM-RFL}), 
where the server can be one of the clients or an independent instance. 
The global ML model with parameters $\Theta$ is partitioned into local models that are stored within individual clients. The server performs global server-side computation and coordinates the computation across clients, while clients perform client-side computation with local model updates.

Moreover, \sys is a \emph{generic} RFL framework in the sense that it allows for integrating with various genres of learning algorithm. We demonstrate this by illustrating how to integrate \sys with two commonly used algorithms, SGD and ADMM.

\begin{figure*}[h!]
\centering
\includegraphics[width=\textwidth]{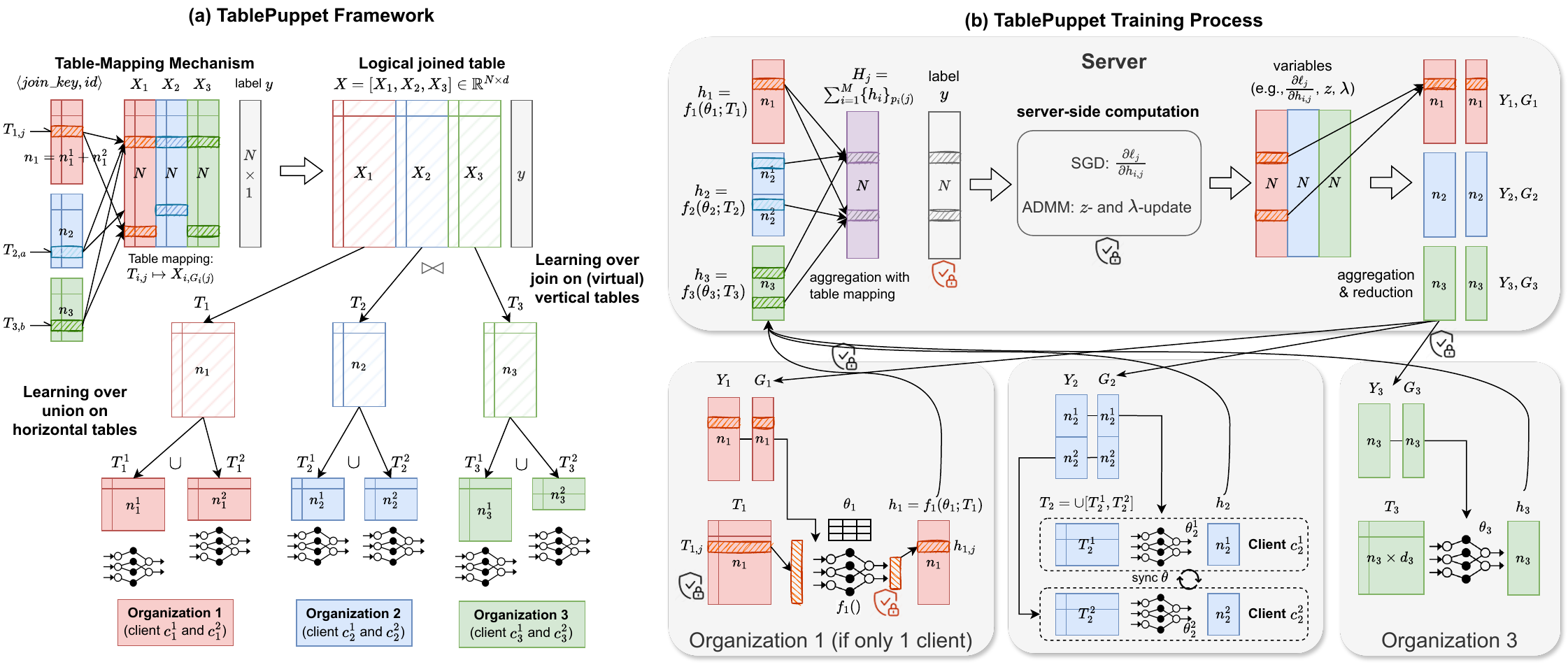}
\vspace{-1.5em}
\caption{\sys framework and its training process. The red (black) locks indicate locations where DP is (can be) applied.}
\label{fig:framework}
\vspace{-0.5em}
\end{figure*}

\begin{table}
\small
\caption{Table of Notations.}
\label{tab:algo}
\vspace{-1em}
\begin{tabular}{|c|l|}
\hline
\textbf{Notation}  & \textbf{Definition}  \\ \hline
$T_i, T_i^q$ &  Organization $i$ owns $T_i$ and its $q$-th client owns $T_i^q$ \\ \hline
$X_i$ & The i-th vertical slice of the joined table $X$ \\ \hline
$X_{i,j} \mapsto T_{i, p_i(j)}$ & The $j$-th tuple of $X_i$ is from the  $p_i(j)$-th tuple of $T_i$ \\ \hline
$h_{i,j} = f_i(\theta_i; x)$ &  $h_{i,j}$ is the prediction of model $f_i(\theta_i)$ on tuple $x$   \\ \hline
\end{tabular}

\end{table}

\subsection{Learning over Join on Vertical Tables (LoJ)} \label{sec:vector-operator}
LoJ first uses a
table mapping mechanism to build a \emph{logical} joined table $X$ to represent the UCQ result without table sharing among clients. It then decomposes/pushes learning over the joined table to each vertical table $T_i$. It finally performs computation and communication reduction for duplicate tuples introduced by joins. 

\subsubsection{\underline{\textbf{Table-Mapping Mechanism}}} 
\label{sec:table-index-mapping}

To represent the global table of UCQ results, our key idea is to join the $\langle \textit{join\_key}, \textit{row\_id}\rangle$ columns of each table to get an index mapping between the logical joined table $X$ and each vertical table $T_i$ as $X_{i,j} \mapsto T_{i,p_i(j)}$, i.e., the $j$-th tuple of $X_{i}$ comes from the $p_i(j)$-th tuple of $T_i$ as shown in Figure~\ref{fig:framework}(a), and $p_i$ denotes the mapping function for $T_i$. 
This index mapping can help us transform FL on the joined table $X$ to FL on each vertical table $T_i$.

\textbf{Step 1: } To obtain this mapping, each client first extracts the $\langle \textit{join\_key}, \textit{row\_id}\rangle$ columns from its table, and then sends these columns to the server. The clients that own labels need to send the labels to the server as well. 
We will detail the related privacy guarantees in Section~\ref{sec:privacy}.

\textbf{Step 2: } After collecting $\langle \textit{join\_key}, \textit{row\_id}\rangle$ columns from all the clients, the server first merges the collected $\langle \textit{join\_key}, \textit{row\_id}\rangle$ of horizontal tables to be $\langle \textit{join\_key}, \textit{row\_id}\rangle$ of vertical table $T_i$. It then joins the \textit{join\_key} columns specified in the UCQ and obtains the mapping $p_i = \left[X_{i,j} \mapsto T_{i, p_i(j)}\right]$ for mapping $X_i$ to $T_i$. 

After that, the computation on each tuple of the joined table $X$ (i.e., $X_{i,j}$) can be transferred to the computation on the corresponding tuple of the vertical table $T_i$, i.e., $T_{i,p_i(j)}$. The server also aggregates the received labels as $y$ based on the table mapping.

\subsubsection{\underline{\textbf{Problem Decomposition and Push Down}}}

LoJ aims to push ML training on the joined table $X$ down to each vertical table $T_i$. Inspired by the vertical FL solution~\cite{VFL-survey}, \sys first decouples the global model to local models by partitioning global model parameters  $\Theta$ to  $\Theta = [\theta_1, \dots, \theta_M]$, where $\theta_i$ is the parameter of the local model $f_i$ associated with $X_i$. Then, \sys transforms the RFL problem of Eq.~\ref{Eq:HybridFL-problem} to an optimization problem on the model prediction of each local model as Eq.~\ref{Eq:FL-over-joins-global}, where $h_{i,j}$ denotes the model prediction of $f_i$ on tuple $X_{i,j}, \forall j \in [N]$:
\begin{gather} 
\min_{\{\theta_i\}_{i=1}^{M}}  \frac{1}{N}  \sum_{j=1}^N \ell \left(\sum_{i=1}^{M} h_{i,j};y_j \right) + \beta\sum_{i=1}^{M} \mathcal{R}_i(\theta_i) \label{Eq:FL-over-joins-global}, \\ 
\text{where } h_{i,j} = f_i(\theta_i; X_{i,j}) = f_i(\theta_i; T_{i, p_i(j)}),  T_i = \cup[T_i^1, \dots, T_i^{Q_i}]. \nonumber 
\end{gather}
Finally, we leverage the table-mapping mechanism to push the learning problem on $X_i$ to the corresponding vertical table $T_i$ with $h_{i,j} = f_i(\theta_i; T_{i, p_i(j)}), \forall j \in [N]$, \emph{which does not require $X_i$ any more.} 
In general, we can use a linear function of $\psi$ to aggregate the model predictions in the loss function $\ell$ as $\ell(\psi (h_{i,j}, \dots, h_{M,j});y_j)$.
In this paper, we use summation for $\psi$ as $\ell(\psi(h_{i,j}, \dots, h_{M,j});y_j) = \ell (\sum_{i=1}^M h_{i,j};y_j )$; we leave the extension to a general linear function in future work. 
For linear models, $\theta_i$ is a matrix as $\theta_i \in \mathbb{R}^{d_i \times d_c}$, where $d_c$ is the number of classes. For deep learning models, $\theta_i$ refers to the parameters of the neural network.

We cannot directly solve the optimization problem of Eq.~\ref{Eq:FL-over-joins-global}, since the server cannot directly compute its gradient $\nabla_{\theta_i} \ell_j = \frac{\partial \ell_j}{\partial \theta_i}$ (here, $ \ell_j = \ell (\sum_{i=1}^{M} h_{i,j};y_j )$), 
which requires that both the local models and tables only reside in individual clients.
To address this, we decompose this optimization problem into sub-problems and push each sub-problem to the corresponding vertical table $T_i$.

Specifically, we discuss below how to push SGD and ADMM down to vertical tables, alongside necessary computation on the server side. For SGD, we focus on how to break down the gradient computation across vertical tables as well as the server. For ADMM, we focus on how to decompose the global optimization problem into separate and independent optimization problems, each associated with a vertical table. Compared to decomposed gradient computation in SGD, these independent optimization problems in ADMM enable more computation on the client side, thereby reducing the  communication rounds with the server:

\vspace{0.2em}
\textbf{(1) SGD.} \label{sec:problem-decomp-sgd}
We can decompose the optimization problem of Eq.~\ref{Eq:FL-over-joins-global} into two sub-problems, by the chain rule $\nabla_{\theta_i} \ell_j =  \frac{\partial \ell_j}{\partial h_{i,j}}\frac{\partial h_{i,j}}{\partial \theta_i}$. 
The first sub-problem is to compute the partial derivative of $\ell_j$ w.r.t. $h_{i,j}$ as $\frac{\partial \ell_j}{\partial h_{i,j}}$ on the server side, since the server can obtain all the model predictions $h_{i,j}$ 
from clients. 
The second sub-problem is to compute the partial derivative of $h_{i,j}$ w.r.t. $\theta_i$ as $\frac{\partial h_{i,j}}{\partial \theta_i}$ on the client side.  Consequently, the client with $T_i$ can compute $\nabla_{\theta_i} \ell_j$
after receiving the $\frac{\partial \ell_j}{\partial h_{i,j}}$ from the server, and further compute the gradient of Eq.~\ref{Eq:FL-over-joins-global} as $\nabla_{\theta_i} F(T_i)$ using Eq.~\ref{Eq:sgd-client-side-vertical1} shown in Table~\ref{tab:push-sgd-admm}. Finally, the client with $T_i$ can update its local $\theta_i$ using Eq.~\ref{Eq:sgd-client-side-vertical2} with learning rate $\eta$. 
Now we have pushed the LoJ problem down to each vertical $T_i$ associated with client-side computation of Eq.~\ref{Eq:sgd-client-side-vertical1} and Eq.~\ref{Eq:sgd-client-side-vertical2}, alongside server-side computation of Eq.~\ref{Eq:SGD-vertical}.  
We can also use mini-batch SGD to perform these equations, by randomly subsampling a batch $B$ from $N$ and change $N$ to $B$ in each equation. However, the client-side computation complexity remains $\mathcal{O}(N)$. 

\vspace{0.3em}
\textbf{(2) ADMM.} \label{sec:problem-decomp}
Unlike SGD, ADMM aims to decompose the global optimization problem of Eq.~\ref{Eq:FL-over-joins-global} into multiple independent optimization sub-problems.
To achieve this, 
we follow the \emph{sharing ADMM} paradigm~\cite{Boyd} to rewrite Eq.~\ref{Eq:FL-over-joins-global} into the Eq.~\ref{Eq:admm-problem} below, by introducing auxiliary variables $z = \{z_j\}_{j=1}^{N} $ where $ z_j \in \mathbb{R}^{d_c}$: 
\begin{align} \begin{matrix} \label{Eq:admm-problem}
 \text{minimize  } & \frac{1}{N}\sum_{j=1}^{N} \ell \left(z_j; y_j \right) + \beta \sum_{i=1}^{M} \mathcal{R}_i(\theta_i), \\
  \text{subject to} &  \sum_{i=1}^{M} h_{i,j}  - z_j = 0, \,\,h_{i,j} = f_i(\theta_i; T_{i, p_i(j)}). \end{matrix}
\end{align}

We then add a quadratic term to the Lagrangian of Eq.~\ref{Eq:admm-problem}, 
known as \textit{augmented Lagrangian}~\cite{Boyd}. After that, we can solve the optimization problem using ADMM with three update steps, including the \textbf{server-side} $z$-update and  $\lambda$-update (Eq.~\ref{Eq:z-update-unoptimized} and~\ref{Eq:lambda-update-unoptimized}), as well as the \textbf{client-side} $\theta$-update (Eq.~\ref{Eq:theta-update-unoptimized}),  as shown in Table~\ref{tab:push-sgd-admm}.
In these equations, the form of $a^t$ refers to the value of $a$ in the $t$-th epoch, while $\rho \in \mathbb{R}$ is a penalty parameter that can be tuned. 
$\{\lambda_j\}_{j=1}^N$ are dual variables and $\lambda_j \in \mathbb{R}^{d_c}$.
The $\{s_{i, j}\}_{j = 1}^{N}$ are residual variables  for each table $T_i$, where  $s_{i, j} = \sum_{k=1, k \neq i}^{M} f_i(\theta_k; T_{k, p_k(j)} ) - z_{j}$.
This decomposition pushes independent optimization problems (i.e., $\theta_i$-update) to each vertical table $T_i$, which requires less frequent communication with the server (only after the $\theta_i$-update finishes) compared to the (mini-batch) SGD.

\begin{table*}
\small
  \centering
  \caption{The problem decomposition for LoJ, using SGD and ADMM with computation/communication reduction.}
  \label{tab:push-sgd-admm}
  \vspace{-1em}
  \begin{tabular}{|c|c|c|}
     \hline
   \textbf{Algorithm} & \textbf{Server} & \textbf{Client with $T_i$}  \\
     \hline
    SGD   & $\begin{aligned} 
     \frac{\partial \ell_j}{\partial h_{i,j}}, \quad \forall j \in [N].   \refstepcounter{equation}\label{Eq:SGD-vertical}(\theequation) \end{aligned}$ &  $\begin{aligned}\nabla_{\theta_i} F(T_i) &= \frac{1}{N} \sum_{j=1}^N \left( \frac{\partial \ell_j}{\partial \theta_i} + \beta \frac{\partial \mathcal{R}_i}{\partial \theta_i} \right), \refstepcounter{equation}\label{Eq:sgd-client-side-vertical1}(\theequation) \\  \theta_i &:= \theta_i - \eta \nabla_{\theta_i} F(T_i).  \refstepcounter{equation}\label{Eq:sgd-client-side-vertical2}(\theequation)\end{aligned}$ \\    \hline
   
      ADMM & $\begin{aligned}  z_j^{t}  &=  \argmin_{z_j} \left( \ell \left( z_j; y_j \right)  -\left(\lambda_j^{t-1}\right)^ \top  z_{j} + \frac{\rho}{2}  {\left \| \sum_{i=1}^{M} f_i(\theta_i^{t}; T_{i, p_i(j)})  - z_{j}   \right\|}^2 \right), \refstepcounter{equation}\label{Eq:z-update-unoptimized}(\theequation) 
      \\
      \lambda_j^{t}  &= \lambda_j^{t-1} + \rho \left(\sum_{i=1}^{M} f_i(\theta_i^{t}; T_{i, p_i(j)}) - z_{j}^{t} \right). \refstepcounter{equation}\label{Eq:lambda-update-unoptimized}(\theequation) 
      \end{aligned}$  & 
      $\begin{aligned} &\theta_i^{t+1}  =  \argmin_{\theta_i}  \left( \beta\mathcal{R}_i (\theta_i) +   \frac{1}{N} \sum_{j=1}^{N} \lambda_j^{t \top} f_i(\theta_i; T_{i, p_i(j)}) \right. \nonumber \\ 
&  \quad \quad\quad \quad\quad\quad\,\,\,\left. +  \frac{1}{N} \sum_{j=1}^{N} \frac{\rho}{2}  {\left \|  s_{i,j}^t + f_i(\theta_i; T_{i, p_i(j)})   \right \|}^2   \right). \refstepcounter{equation}\label{Eq:theta-update-unoptimized}(\theequation) 
      \end{aligned}$
      
      \\    \hline 
       SGD-Opt &   $\begin{aligned} 
Y_{i,j} &= \sum_{g \in G_i(j)} \frac{\partial \ell_g}{\partial h_{i,g}},  \quad \forall j \in [n_i],  \refstepcounter{equation}\label{Eq:SGD-Y-formula}(\theequation)  \\
G_{i,j} &= \left| G_i(j) \right|,  \quad  \forall j \in [n_i]   \refstepcounter{equation}. \label{Eq:SGD-G-formula}(\theequation) 
\end{aligned}$ &  
$\begin{aligned}\nabla_{\theta_i} F(T_i) &= \frac{1}{N} \sum_{j=1}^{n_i} \left( Y_{i,j} \frac{\partial h_{i,j}}{\partial \theta_i} + \beta G_{i,j} \frac{\partial \mathcal{R}_i}{\partial \theta_i} \right), \refstepcounter{equation}\label{Eq:sgd-client-side-vertical1-opt}(\theequation) \\  \theta_i &:= \theta_i - \eta \nabla_{\theta_i} F(T_i).  \refstepcounter{equation}\label{Eq:sgd-client-side-vertical2-opt}(\theequation)\end{aligned}$
\\    \hline
    ADMM-Opt & $\begin{aligned} & \text{Apart from Eq.~\ref{Eq:z-update-unoptimized} and \ref{Eq:lambda-update-unoptimized}, the server also performs:} \\
Y_{i,j}^t &= \sum_{g \in G_i(j)} (\lambda_g^{t} + \rho s_{i,g}^{t}),  \quad \forall j \in [n_i],  \refstepcounter{equation}\label{Eq:App-Y-formula}(\theequation)  \\
G_{i,j} &= \left| G_i(j) \right|,  \quad  \forall j \in [n_i]   \refstepcounter{equation}. \label{Eq:App-G-formula}(\theequation) 
\end{aligned}$ & $\begin{aligned}
&\theta_i^{t+1} =  \argmin_{\theta_i} \left( \beta\mathcal{R}_i (\theta_i) +   \frac{1}{N}  \sum_{j=1}^{n_i} ({Y_{i,j}^t})^\top  f_i(\theta_i; T_{i,j})  \right. \nonumber \\
&\quad \quad\quad \quad\quad\quad\quad\quad\,\,\,\left.+ \frac{1}{N}  \sum_{j=1}^{n_i}\frac{\rho G_{i,j}}{2}  \left \| f_i(\theta_i; T_{i,j}) \right \|^2 \right). \refstepcounter{equation}\label{Eq:theta-Y-G-update-red}(\theequation)
\end{aligned}$ \\    \hline
  \end{tabular}
  
\end{table*}

\begin{table*}
\small
  \centering
  \caption{The problem decomposition for LoU, using optimized SGD and ADMM.}
  \vspace{-1em}
  \label{tab:push-sgd-admm-h}
  \begin{tabular}{|c|c|c|}
     \hline
   \textbf{Algorithm} & \textbf{Coordinator (Server)} & \textbf{Client with $T_i^q$}  \\
     \hline
    SGD-Opt   & $\begin{aligned} 
     \nabla_{\theta_i} F(T_i) = \frac{1}{N}\sum_{q=1}^{Q_i} \nabla_{\theta_i} F(T_i^q).   \refstepcounter{equation}\label{Eq:SGD-horizontal}(\theequation) \end{aligned}$ &  $\begin{aligned}\nabla_{\theta_i} F(T_i^q) &= \sum_{j=1}^{n_i^q} \left( Y_{i,j} \frac{\partial h_{i,j}}{\partial \theta_i} + \beta G_{i,j} \frac{\partial \mathcal{R}_i}{\partial \theta_i} \right), \refstepcounter{equation}\label{Eq:sgd-client-side-horizontal1}(\theequation) \\  \theta_i &:= \theta_i - \eta \nabla_{\theta_i} F(T_i).  \refstepcounter{equation}\label{Eq:sgd-client-side-horizontal3}(\theequation)\end{aligned}$ \\    \hline

      ADMM-Opt & $\begin{aligned}   w_i^{\tau} & =  \argmin_{w_i} \left( \beta\mathcal{R}_i (w_i) + \frac{M \rho}{2} {\left \| w_i - \overline{\theta_i^{\tau}} - \overline{u_i^{\tau-1}}\right \|}^2 \right), \refstepcounter{equation}\label{Eq:ADMM-horizontal-w-update}(\theequation) 
      \\
      u_i^{q,\tau} & = u_i^{q,\tau-1} + \theta_i^{q,\tau} - w_i^{\tau}. \refstepcounter{equation}\label{Eq:ADMM-horizontal-u-update}(\theequation) 
      \end{aligned}$  & 
      $\begin{aligned} &\theta_i^{q, \tau+1} =  \argmin_{\theta_i^{q}} \left(\frac{1}{N} l\left(\theta_i^{q};  T_i^{q} \right) + \frac{\rho}{2} {\left \| \theta_i^{q} - w_i^{\tau} + u_i^{q,\tau} \right \|}^2 \right). \refstepcounter{equation}\label{Eq:ADMM-horizontal-theta-update}(\theequation) 
      \end{aligned}$
      \\    \hline 
  \end{tabular}
    
\end{table*}

\subsubsection{\underline{\textbf{Computation and Communication Reduction}}}
\label{sec:compute-red}

Though ADMM can reduce the communication rounds compared to SGD, its computation and communication complexity remains $\mathcal{O}(N)$ as that of SGD. Regarding the computation complexity, both SGD and ADMM need to perform $\sum_{j=1}^{N}(\cdot)$ during client-side computation using Eq.~\ref{Eq:sgd-client-side-vertical1} and ~\ref{Eq:theta-update-unoptimized}.
Here, $N$ refers to the length of the joined table $X$, which can be orders of magnitude larger than the length of each vertical table $T_i$ as $n_i$, due to the duplicate tuples introduced by joins. For example, if both $X_{i,a}$ and $X_{i,b}$ are generated by the same tuple $T_{i, k}$ after the join, $T_{i, k}$ will be visited twice during the client-side computation of SGD and ADMM. 
Regarding the communication complexity, the server needs to send its computation results to each client with $T_i$ in each epoch, such as the partial gradient $\frac{\partial \ell_j}{\partial h_{i,j}}$ of SGD and the $s_{i,j}$, $\lambda_j$ of ADMM. These variables have the length of $N$, leading to communication complexity of $\mathcal{O}(N)$.

To minimize the computation and communication overhead, our LoJ step conducts the following two optimization strategies: (1) reduce the client-side computation on duplicate tuples by aggregating the server-side computation results (variables); and (2) communicate the aggregated variables instead of the original ones between the server and clients. Below we detail how to perform them for SGD and ADMM:

\vspace{0.5em}
\textbf{(1) SGD.} After reviewing the chain rule of $\nabla_{\theta_i} \ell_j =  \frac{\partial \ell_j}{\partial h_{i,j}}\frac{\partial h_{i,j}}{\partial \theta_i}$, we found that 
the second part $\frac{\partial h_{i,j}}{\partial \theta_i}$ is the same for duplicate tuples from $T_i$, because this part is only determined by the model function $f_i$ and each tuple of $T_{i}$. In this case, we can aggregate the first part for duplicate tuples on the server side, and then send the aggregated variables to the clients for reducing the client-side computation.

To facilitate the aggregation on the server side, we first construct a \emph{reverse} table mapping $T_{i, j} \mapsto X_{i, G_i(j)}$, which means $T_{i, j}$, i.e., the $j$-th tuple of $T_i$, is mapped to multiple tuples in $X_i$, denoted as $\{X_{i, g}\}_{g \in G_i(j)}$.
We then
aggregate the $\frac{\partial \ell_j}{\partial h_{i,j}}$ of duplicate tuples as shown in Eq.~\ref{Eq:SGD-Y-formula} and Eq.~\ref{Eq:SGD-G-formula}. Using the aggregated variables $Y_{i,j}$ and $G_{i,j}$, we can rewrite the client-side computation of Eq.~\ref{Eq:sgd-client-side-vertical1} into Eq.~\ref{Eq:sgd-client-side-vertical1-opt}, where ${G}_i$ remains the same in each epoch.

Note that Eq.~\ref{Eq:sgd-client-side-vertical1-opt} has transformed the computation of $\sum_{j=1}^{N}(\cdot)$ to  $\sum_{j=1}^{n_i}(\cdot)$
and therefore the computation overhead is reduced from $O(N)$ to $O(n_i)$.
Moreover, the server can send ${Y}_i \in \mathbb{R}^{n_i \times d_c}$ and $G_i \in \mathbb{R}^{n_i}$ to the client that owns $T_i$, instead of sending $\frac{\partial \ell_j}{\partial h_{i,j}} \in  \mathbb{R}^{N \times d_c}$ and the table-mapping that are in $O(N)$. 
Therefore, the communication overhead between the server and the clients is reduced from $\mathcal{O}(N)$ to $\mathcal{O}(n_i)$ as well. However, for mini-batch SGD with batch size of $B$, the server needs to perform the aggregation for the duplicate tuples inside each batch, so the reduced computation and communication is between $\mathcal{O}(n_i)$ and $\mathcal{O}(N)$.

\vspace{0.5em}
\textbf{(2) ADMM.} For the client-side computation of Eq.~\ref{Eq:theta-update-unoptimized}, duplicate tuples will lead to duplicate computation of $f_i(\theta_i; T_{i, p_i(j)})$. To reduce this computation, we can perform the optimization strategy similar to SGD. The server combines and aggregates the variables (e.g., $\lambda_j$ and $s_{i,j}$) of duplicate tuples as shown in Eq.~\ref{Eq:App-Y-formula} and Eq.~\ref{Eq:App-G-formula}. Using $Y_{i,j}$ and $G_{i,j}$, we can rewrite the $\theta_i$-update as Eq.~\ref{Eq:theta-Y-G-update-red}.

Similar with the optimization on SGD, Eq.~\ref{Eq:theta-Y-G-update-red} transforms the computation of $\sum_{j=1}^{N}(\cdot)$ to  $\sum_{j=1}^{n_i}(\cdot)$
and thus reduces the computation overhead from $O(N)$ to $O(n_i)$.
Moreover, the server can send ${Y}_i$ and $G_i$ to the client that owns $T_i$, instead of sending $\lambda$, $s_{i}$ and the table-mapping that are in $O(N)$ as shown by Eq.~\ref{Eq:theta-update-unoptimized}. 
Therefore, the communication overhead  is also reduced from $\mathcal{O}(N)$ to $\mathcal{O}(n_i)$. More details can be found in the Appendix~\cite{Appendix}.

\subsection{Learning over Union on Horizontal Tables} \label{sec:horizontal}

LoJ has pushed ML training down to each vertical table $T_i$. 
The next question is how to push the computation down to each horizontal table $T_i^q$. 
For SGD, we can decompose the gradient computation and synchronize it for local model update, as the gradient computation on each tuple is independent. For ADMM, to decompose the optimization problem of $\theta_i$-update, we can also use SGD. However, to reduce the number of communication rounds, we use horizontal ADMM instead to decompose it to independent optimization problems on horizontal tables:

\vspace{0.5em}
\textbf{(1) SGD.} Our goal is to push the computation on $T_i$, i.e., Eq.~\ref{Eq:sgd-client-side-vertical1-opt} and Eq.~\ref{Eq:sgd-client-side-vertical2-opt}, down to the clients that own $\{T_i^q\}_{q \in [Q_i]}$. Our key idea is to decouple the gradient computation 
while performing the same model update 
in each client as that of Eq.~\ref{Eq:sgd-client-side-vertical2-opt}. First, we allocate the same model with the same $\theta_i$ for each client with $T_i^q$.
Second, we decompose the gradient computation of Eq.~\ref{Eq:sgd-client-side-vertical1-opt} into sub-problems as follows:
(1) each client first performs partial gradient computation on $T_i^q$ as Eq.~\ref{Eq:sgd-client-side-horizontal1}; (2) a coordinator of the clients then aggregates the partial gradients as Eq.~\ref{Eq:SGD-horizontal} and sends the aggregate back to the clients; (3) each client updates the model using Eq.~\ref{Eq:sgd-client-side-horizontal3}. In particular, the server can act as the coordinator as well.

\textbf{(2) ADMM.}
For ADMM, we decompose an independent optimization problem, i.e., the $\theta_i$-update, into sub-problems w.r.t. the $T_i^q$'s, by leveraging \textit{consensus ADMM}~\cite{Boyd}.
Conceptually, this decomposition ``pushes'' ADMM through the \emph{union} operation down to the horizontal tables.
Specifically, we first rewrite Eq.~\ref{Eq:theta-Y-G-update-red} as Eq.~\ref{Eq:min-theta-update-with-q}:
\begin{align}
\begin{matrix} \label{Eq:min-theta-update-with-q}
    \text{minimize}    \frac{1}{N}\sum_{q=1}^{Q_i}  l\left(\theta_i^{q}; T_i^{q} \right) + \beta\mathcal{R}_i (\theta_i),  \\
      l\left(\theta_i^{q}; T_i^{q}\right) = \sum_{j=1}^{n_i^q} \left[ (Y_{i,j}^{q,t})^\top f_i(\theta_i^q; T_{i, j}^q)  + \frac{\rho G_{i,j}^q }{2}   
                {\left \|f_i(\theta_i^q; T_{i, j}^q) \right \|}^2  \right]
\end{matrix}
\end{align}
Here, $\theta_i^q$ refers to the model parameters w.r.t. $T_i^q$. $T_{i, j}^q$ denotes the $j$-th tuple in $T_i^q$, and $Y_{i,j}^{q,t}$ refers to the $j$-th element of the $q$-th part of $Y_{i}$ in the $t$-th epoch.
We then rewrite Eq.~\ref{Eq:min-theta-update-with-q} as Eq.~\ref{Eq:horizontal-admm-problem} by introducing auxiliary variables $w_i$ to approximate each $\theta_i^q$:

\begin{align}
\begin{matrix} \label{Eq:horizontal-admm-problem}
 \text{minimize  } &  \sum_{q=1}^{Q_i} \frac{1}{N} l\left(\theta_i^{q}; T_i^{q} \right)  + \beta\mathcal{R}_i (w_i),\\
 \text{subject to} &  \theta_i^{q} - w_i = 0, \quad \forall q \in [Q_i].
\end{matrix}
\end{align}
We can now use consensus ADMM to solve this optimization problem with three update steps (Table~\ref{tab:push-sgd-admm-h}).
Here, $\overline{\theta_i^{\tau}}$ is the average of $\{\theta_i^{q,\tau}\}_{q \in [Q_i]}$ and $\overline{u_i^{\tau-1}}$ is the average of $\{u_i^{q,\tau-1}\}_{q \in [Q_i]}$, where $u_i$ is the scaled dual variable and $\tau$ denotes the $\tau$-th epoch of the horizontal ADMM. 
The coordinator performs $w$-update and the $u$-update, while each client $q$ that owns $T_i^q$ can perform the $\theta_i^q$-update. 
During each epoch of the horizontal ADMM, each client $q$ sends its updated $\theta_i^{q}$ to the coordinator and the coordinator returns $w_i$ and $u_i^q$ to the client $q$ for model update with one round of communication.

\section{Implementation of \sys}
\label{sec:ADMM-RFL}

\begin{figure}
\centering
\includegraphics[width=\columnwidth]{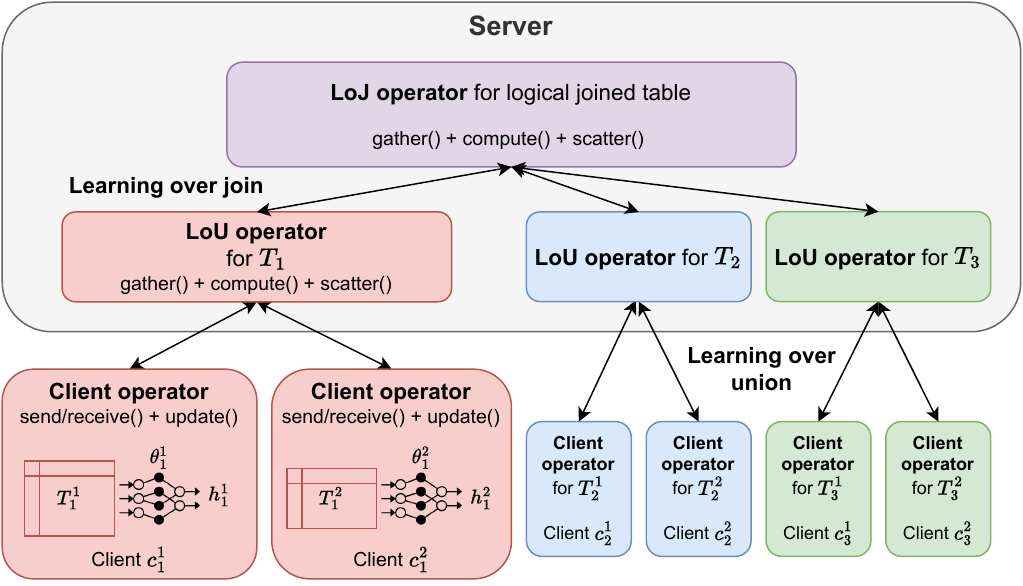}
\vspace{-1em}
\caption{The architecture of \sys.}
\label{architecture}

\end{figure}

\sys adopts a server-client architecture (Figure~\ref{architecture}), 
where the global ML model with parameters $\Theta$ is partitioned into local models with $\{\theta_i^q\}_{i \in [M], q \in [Q_i]}$ stored in clients.
The server and clients collaboratively and iteratively train these local models. We abstract the training process as an \emph{execution plan} of three \emph{physical operators} (Section~\ref{sec:training-process}), which unifies the computation and communication of SGD/ADMM (Section~\ref{sec:algorithms}). 
\sys also ensures data privacy for \emph{both} features and labels with differential privacy (Section~\ref{sec:privacy}).

\vspace{-0.5em}
\subsection{Training Process with Physical Operators} \label{sec:training-process}

We illustrate the training process in Algorithm~\ref{algo:training-process}.
which mainly consists of two loops: (1) the outer loop that conducts \emph{learning over join} (LoJ) and (2) the inner loop that performs \emph{learning over union} (LoU). To implement and unify the training processes of SGD and ADMM, 
we design three physical operators that abstract the computation and communication of \sys. 
The three operators include a \emph{LoJ operator}, a \emph{LoU operator}, and a \emph{client operator} for client-side model updates. 
As shown in Table~\ref{tab:operators}, each operator contains three functions, including a $\emph{compute}()$  for the computation, as well as $\emph{gather/scatter}()$ or $\emph{send/receive}()$ for the communication. 

\begin{algorithm}[t!]
\footnotesize
\DontPrintSemicolon

\KwIn{ 
Suppose $X = \bowtie[T_1, \dots, T_M]$ and $T_i = \cup[T_i^1, T_i^2, \dots, T_i^{Q_i}]$. There are one server and $Q$ clients with $Q=\sum_{i=1}^{M}Q_i$. Each client owns a local model and a table as $T_i^q$. The number of communication rounds for outer loop is $\mathcal{T}$ and inner loop is $\mathcal{T}'$. For SGD, $\mathcal{T} = KN/B$ and $\mathcal{T}' = 1$, where $K$ is the epoch number, $N$ is the tuple number of $X$, and $B$ is the batch size. For ADMM, $\mathcal{T} = K$.
}

Server performs \textbf{table mapping} (Section~\ref{sec:table-index-mapping}). \;

\For(\quad\Comment*[h]{\textcolor{blue}{Outer loop}}) {communication round  $ t \in [\mathcal{T}]$} { 
    
    \For {each client with $T_i^q$, $i \in [M], q \in [Q_i]$} {
        
        \textbf{computes} model predictions and \textbf{send} them to the server.

    }
    Server \textbf{gathers} and \textbf{aggregates} model predictions. \; 
    Server \textbf{does} server-side computation (SGD-/ADMM-Opt in Table~\ref{tab:push-sgd-admm}).\;
    Server \textbf{scatters} computation results to the clients. \;

    \For(\quad\Comment*[h]{\textcolor{blue}{Inner loop}})  {$i \in [M]$} {
        \For (\quad\Comment*[h]{$\mathcal{T}' = 1$ if $Q_i=1$}){communication round $ \tau \in [\mathcal{T}']$ }  { 
            \If {$Q_i > 1$} {
                \For {each client with $T_i^q$, $q \in [Q_i]$} {
                    \textbf{computes} and \textbf{sends} variables (SGD: Eq.~\ref{Eq:sgd-client-side-horizontal1} with \textbf{DP}, ADMM: local model parameters) to the server.\; 
                }
                Server \textbf{performs} coordinator-side computation (SGD: Eq.~\ref{Eq:SGD-horizontal}, ADMM: Eq.~\ref{Eq:ADMM-horizontal-w-update} and \ref{Eq:ADMM-horizontal-u-update}).\;
                Server \textbf{scatters} computation results to the clients. \;
            }
            \For {each client with $T_i^q$, $q \in [Q_i]$} {
                \textbf{performs} client-side computation and \textbf{model update} (SGD: Eq.~\ref{Eq:sgd-client-side-horizontal3}, ADMM: Eq.~\ref{Eq:ADMM-horizontal-theta-update} with \textbf{DP}). \;
            }
        }
    }
    
}

 \caption{\sys training process.  } \label{algo:training-process}
\end{algorithm}

\textbf{(1) LoJ operator:} This operator is performed in the server for LoJ, with the assumption that each LoU operator represents a vertical table $T_i$. It first uses $\textit{gather}()$ to collect model predictions from each LoU operator, and then aggregates these model predictions based on the table mapping. After that, it performs server-side computation in $\textit{compute}()$. Finally, it performs $\textit{scatter}()$ to distribute the computation results to each LoU operator, as if the computation results were scattered to each $T_i$. As shown in Algorithm~\ref{algo:training-process}, the LoJ operator is responsible for the steps 5-7 of the training process.

\textbf{(2) LoU operator:} This operator can be performed in the server (or in each organization) as the coordinator of the clients with horizontal tables $\{T_i^q\}_{q \in [Q_i]}$. In the outer loop, it uses $\textit{gather}()$ to collect model predictions from clients in the same organization, and combines them as model predictions on a vertical $T_i$, which are sent to the LoJ operator. It also applies $\emph{scatter}()$ to distribute computation results from the LoJ operator to the clients. In the inner loop, it gathers partial gradients or variables from clients using $\textit{gather}()$, performs coordinator-side computation using $\textit{compute}()$, and distributes computation results back to clients with $\emph{scatter}()$. As shown in Algorithm~\ref{algo:training-process}, the LoU operator is responsible for the steps 13-14 of the training process.

\textbf{(3) Client operator:} This operator is performed in each client with horizontal table $T_i^q$.
It applies $\emph{send}()$ to transmit model predictions, partial gradients, or model parameters to the LoU operator (coordinator) for aggregation and computation. 
It gets computation results, such as aggregated gradients and variables, from the LoU operator using $\emph{receive}()$. 
Finally, it performs $\textit{compute}()$ for model update. 
As shown in Algorithm~\ref{algo:training-process}, the client operator is responsible for the steps 4, 12, and 16 of the training process.

\vspace{3pt}
\subsection{SGD/ADMM with \sys} \label{sec:algorithms}

By combining the three physical operators in various ways and adjusting the implementations of their inner functions, we can implement SGD/ADMM not only for RFL but also for FL atop the vertical tables being joined (without the horizontal partitions).
For ease of exposition, we call this special case \textbf{RFL-V} in the rest of this paper.
As shown in Table~\ref{tab:algo}, for RFL-V where each organization only has one client, we can just combine the LoJ operator and the client operator to implement \textbf{RFL-SGD-V} and \textbf{RFL-ADMM-V}. 
For RFL, we use all three operators with functions shown in Table~\ref{tab:operators} to implement \textbf{RFL-SGD} and \textbf{RFL-ADMM}.

\begin{table}
\small
\caption{The physical operators with computation and communication functions, which cover both SGD and ADMM.}
\label{tab:operators}
\vspace{-1em}
\begin{tabular}{|l|l|l|l|}
\hline
\textbf{Operator}  & \textbf{Function} & \textbf{SGD} & \textbf{ADMM} \\ \hline
\multirow{3}{*}{\shortstack{\textbf{LoJ}\\\textbf{operator}}} & \textit{gather}() & model predictions & model predictions \\ \cline{2-4}
                    & \textit{compute}()  & Eq.~\ref{Eq:SGD-Y-formula},~\ref{Eq:SGD-G-formula}  &  Eq.~\ref{Eq:z-update-unoptimized},~\ref{Eq:lambda-update-unoptimized},~\ref{Eq:App-Y-formula},~\ref{Eq:App-G-formula} \\ \cline{2-4}
                    & \textit{scatter}() & partial derivatives  &  auxiliary variables \\ \hline
\multirow{4}{*}{\shortstack{\textbf{LoU}\\\textbf{operator}}}  & \multirow{2}{*}{\textit{gather}()}  & model predictions,
  &  model predictions, \\ 
  &    & partial gradients
  &  model parameters \\ \cline{2-4}
                    & \textit{compute}()  & Eq.~\ref{Eq:SGD-horizontal}  &  Eq.~\ref{Eq:ADMM-horizontal-w-update} and Eq.~\ref{Eq:ADMM-horizontal-u-update} \\ \cline{2-4}
                    & \textit{scatter}()  & aggregated gradients  &  auxiliary variables \\ \hline
\multirow{4}{*}{\shortstack{\textbf{Client}\\\textbf{operator}}}  & \multirow{2}{*}{\textit{send}()}  & model predictions,  &  model predictions, \\ 
                     &    & partial gradients
                        &  model parameters \\ \cline{2-4}
                    & \textit{receive}()  &  aggregated gradients  &  auxiliary variables \\ \cline{2-4}
                    & \textit{compute}()  & Eq.~\ref{Eq:sgd-client-side-horizontal1} and ~\ref{Eq:sgd-client-side-horizontal3}  &  Eq.~\ref{Eq:ADMM-horizontal-theta-update} \\\hline
\end{tabular}

\end{table}

\begin{table}
\small
\caption{The SGD and ADMM algorithms atop \sys.}
\label{tab:algo}
\vspace{-1em}
\begin{tabular}{|c|l|l|}
\hline
\textbf{Scenario}  & \textbf{Algorithm} & \textbf{Implementation of operators} \\ \hline
\multirow{2}{*}{\textbf{RFL-V}} 
                   & \textbf{RFL-SGD-V} & LoJ: Eq.~\ref{Eq:SGD-Y-formula},~\ref{Eq:SGD-G-formula}, Client: Eq.~\ref{Eq:sgd-client-side-vertical1-opt},~\ref{Eq:sgd-client-side-vertical2-opt} \\ \cline{2-3}
                   & \textbf{RFL-ADMM-V} & LoJ: Eq.~\ref{Eq:z-update-unoptimized},~\ref{Eq:lambda-update-unoptimized}, \ref{Eq:App-Y-formula}, \ref{Eq:App-G-formula}, Client: Eq.~\ref{Eq:theta-Y-G-update-red} \\ \hline
\multirow{2}{*}{\textbf{RFL }} & \textbf{RFL-SGD} & As that in Table~\ref{tab:operators} (SGD column) \\ \cline{2-3}
                   & \textbf{RFL-ADMM} & As that in Table~\ref{tab:operators} (ADMM column) \\ \hline
\end{tabular}

\end{table}

\subsection{Privacy Guarantees} \label{sec:privacy}

\sys introduces \textit{differential privacy} (DP)~\cite{dwork2014algorithmic} to ensure data privacy for both the table features and labels. We adopt the DP definition for RFL as follows. 

\begin{definition} [$(\epsilon,\delta)$-DP~\cite{dwork2014algorithmic}]
\label{def:dp}
A randomized algorithm 
$\mathcal{M}: \mathcal{X}^n \mapsto \Theta$ satisfies $(\epsilon, \delta)$-DP if, for every pair of neighboring datasets (i.e., the joined tables) $X, X^{\prime} \in  \mathcal{X}^n$ that differ by one single tuple,
and every possible (measurable) output set $E \subseteq \Theta $, the following inequality holds: $\operatorname{Pr}[\mathcal{M}(X) \in E] \leq e^{\epsilon} \operatorname{Pr}\left[\mathcal{M}\left(X^{\prime}\right) \in E\right]+\delta$.
\end{definition}
We focus on addressing two key problems: \textbf{(1) where to inject DP noises} and \textbf{(2) how to allocate the privacy budget for RFL}. We present the details of our solutions below.

\vspace{-0.5em}
\paragraph*{Noise Injection}

Possible options for noise injections include the individual tables, local model parameters, or the diverse variables communicated between the server and clients, shown as black/red locks in Figure~\ref{fig:framework}(b). The added noises should ensure the privacy of labels/features as well as SGD/ADMM. Our solution is to use separate DP mechanisms to safeguard labels and features, as labels are sent to the server whereas table features are kept in clients. To achieve label-level DP, we directly \textbf{add noise to the raw labels} (Section \ref{sec:dp_table_mapping}). To protect features, we \textbf{perturb local model parameters} on the client side (Section \ref{sec:dp_model_training}) instead of perturbing the communicated variables. Our rationale for making this design decision is that SGD and ADMM have different communicated variables so it is challenging to add noises to these variables in a uniform way.
Adding noises to the local model parameters can automatically ensure DP of the communicated variables, due to the  post-processing property of DP~\cite{dwork2014algorithmic}.

\vspace{-0.5em}
\paragraph*{Budget Allocation}
RFL includes both LoJ and LoU that are performed on both vertical and horizontal tables and the challenge lies in how to calculate the privacy budget for this complex scenario. We allocate privacy budgets for labels and features separately: (1) For label protection, the budget is determined by the one-time noise added on raw labels (using noisy labels for training does not incur additional privacy costs due to the post-processing property of DP).
(2) For feature protection, our solution is to extend the moments accountant~\cite{abadi2016deep} to our framework and track the budget over multiple local training steps on each table (Section~\ref{sec:privacy-dp}).

\subsubsection{Privacy guarantee for labels} 
\label{sec:dp_table_mapping}

In \sys, 
clients need to add noises to the labels before sending them to the server. 
Specifically,  
as shown by Eq.~\ref{eq:label-dp}, \sys first adds Laplace noise with per-coordinate standard deviation $\lambda$ to the labels $y$, to satisfy the DP guarantee, which results in a perturbed label vector. Then, we identify the class that retains the maximum value in the  perturbed label vector as the new class label, with $N_{\mathrm{class}}$ representing the total number of classes. 
For continuous/numeric labels, $N_{\mathrm{class}} = 1$.

\begin{align}
    \hat {y} \gets \mathrm{OneHot}(y) + \mathrm{Laplace}(\lambda),  \quad 
    \hat {y} \gets \arg\max_{j\in [N_{\mathrm{class}}]} \hat {y}_j. \label{eq:label-dp}
\end{align} 

In addition to the labels, 
clients can use cryptographic methods such as SHA-256 hashing to encrypt the $\textit{join\_key}$. 
The server then uses the encrypted $\textit{join\_key}$ to perform joins and unions, assuming that hash collision is rare. 

\begin{table*}
\small
  \centering
  \caption{The computation and communication complexity comparison, where
  $B$ refers to the \textit{batch size} and $\alpha_i \in [n_i, N]$.
  }
  \label{complexity-analysis-table}
  \vspace{-1em}
  \begin{tabular}{lcccc|cc}
  \toprule
   \textbf{Complexity (Per epoch)} & \textbf{VFL-SGD} & \textbf{VFL-ADMM}  & \textbf{RFL-SGD-V} & \textbf{RFL-ADMM-V}  & \textbf{RFL-SGD}  
   & \textbf{RFL-ADMM}\\
    \midrule
    Computation (Server)   & $  \mathcal{O}\left(N  \right)$  &  $\mathcal{O}\left( N  \right)$ &  $\mathcal{O}\left( N  \right)$ &  $\mathcal{O}\left( N  \right)$   
    & $\mathcal{O}\left(N\right) + \mathcal{O}\left(MN/B\right)$
    &  $\mathcal{O}\left(N\right) + \mathcal{O}\left(M\mathcal{T}'\right)$   \\ 
    Computation (Client $c_i^q$)   & $ \mathcal{O}\left( N  \right)$ &    $\mathcal{O}\left( N  \right)$ &  $\mathcal{O}\left( \alpha_i  \right)$ & $\mathcal{O}\left(n_i \right)$  
    & $\mathcal{O}\left(\alpha_i / |Q_i|\right)$
    & $\mathcal{O}\left( \mathcal{T}' n_i^q \right)$  \\
    \midrule
    Communication rounds   & $  \mathcal{O}\left(N / B \right)$  & $\mathcal{O}\left(1  \right)$ & $\mathcal{O}\left(N / B \right)$  & $\mathcal{O}\left(1  \right)$ 
     &  $\mathcal{O}\left(N/B\right)$
     & $\mathcal{O}\left(\mathcal{T}'  \right)$\\

    Comm. cost (Server $\leftrightarrow$ Clients)   & $  \mathcal{O}\left(MN \right)$  & $\mathcal{O}\left(MN  \right)$ & $\mathcal{O}\left(\sum_{i=1}^M \alpha_i  \right)$  & $\mathcal{O}\left(\sum_{i=1}^M n_i  \right)$ 
    & $\mathcal{O}\left(\sum_{i=1}^M \alpha_i\right)  + \mathcal{O}\left(QN/B\right)$
    & $\mathcal{O}\left(\sum_{i=1}^M n_i + \mathcal{T}' Q\right)$ \\
    \bottomrule
  \end{tabular}
  
\end{table*}

\subsubsection{Privacy guarantee for features} 
\label{sec:dp_model_training}
To protect the communicated variables as well as each client's local data, we introduce DP-SGD (i.e., clipping and perturbing)~\cite{abadi2016deep} when updating each local model so that it satisfies $(\epsilon , \delta)$-DP. 
Since DP holds for any post-processing on top of the data, the communicated variables based on client's local model, such as model predictions, partial derivatives/gradients, and model parameters, also satisfy $(\epsilon, \delta)$-DP.

Specifically, when updating the local model of each client, we first clip per-sample gradient $\mathcal{G}_j$  with $L^2$-norm threshold $C$, and then add Gaussian noise sampled from $\mathcal{N} (0, \sigma^2C^2)$ to the averaged batch gradient as $\mathcal{G} \gets \frac{1}{B}  \left(\sum_{j=1}^{B}\mathtt{Clip}\left(\mathcal{G}_j, \mathcal{C}\right) + \mathcal{N}\left(0, \sigma^{2} \mathcal{C}^{2}\right)\right)$.
Here, $\mathcal{G}_j$ is the gradient of the $j$-th sample/tuple, and $\mathcal{G}$ is the averaged gradient over a batch. $B$ refers to the batch size. 
For SGD, we perturb $\nabla_{\theta_i} F(T_i)$, by performing per-sample gradient $\texttt{Clip}$ in Eq.~\ref{Eq:sgd-client-side-horizontal1} with $\mathcal{G}_j = \frac{\partial \ell}{\partial \theta_i} + \beta \frac{\partial \mathcal{R}_i}{\partial \theta_i}$  and adding noise $\mathcal{N}$.
For ADMM, we directly perform the formula of $\mathcal{G}$  while computing the gradient of the $\theta_i^q$-update optimization problem (Eq.~\ref{Eq:ADMM-horizontal-theta-update}).

\section{Formal Analysis}

\subsection{Complexity analysis} \label{complexity_analysis}

Table~\ref{complexity-analysis-table} summarizes the computation and communication complexity of the SGD/ADMM algorithms atop \sys, in comparison with traditional VFL methods (\textbf{VFL-SGD} and \textbf{VFL-ADMM}), which are hypothetically assumed to be running on the vertical partitions of the entire joined table. 
For simplicity, we omit the complexity analysis of the table-mapping mechanism in Table~\ref{complexity-analysis-table}. 
We also represent the computation/communication complexity in terms of the tuple number (e.g., $N$ and $n_i$), since other parameters, such as feature dimension (e.g., $d_i$), are correlated with the tuple number.

\subsubsection{VFL and RFL-V} Each organization has only one client and the client owns a vertical table. Below we analyze the complexity of \textbf{VFL-SGD}, \textbf{VFL-ADMM}, and our \textbf{RFL-SGD-V}, \textbf{RFL-ADMM-V}.

\textbf{Computation complexity.} The server-side computation complexity is $\mathcal{O}(N)$ per epoch for the four algorithms, since they all perform computation on the mapped model predictions with $N$ tuples. 
The computation complexity of each client is $\mathcal{O}(N)$ for \textbf{VFL-SGD} and \textbf{VFL-ADMM}, since they compute on vertical partitions of the joined table. Our \textbf{RFL-SGD-V} and
\textbf{RFL-ADMM-V} reduces the computation complexity to $\mathcal{O}(\alpha_i)$ and $\mathcal{O}(n_i)$, respectively, where $\alpha_i \in [n_i, N]$ (Section \ref{sec:compute-red}).

\textbf{Communication complexity.} \textbf{VFL-SGD} and \textbf{RFL-SGD-V} split the joined table into multiple batches with $\textit{batch\_size} = B$ and gathers/scatters model predictions/gradients for every batch. 
Therefore, the number of communication rounds is $N/B$ per epoch. The corresponding cost between the server and all clients is $O(MN)$ per epoch for \textbf{VFL-SGD} and is reduced to $\mathcal{O}(\sum_{i=1}^M \alpha_i)$ for \textbf{RFL-SGD-V}. 
In contrast, \textbf{VFL-ADMM} and \textbf{RFL-ADMM-V} decouple the ML training problem into sub-problems and solve each sub-problem in each client. 
Thus, they only need to gather/scatter model predictions/variables \emph{once} per epoch between the server and clients, introducing $\mathcal{O}(1)$ communication rounds. 
However, \textbf{VFL-ADMM} still suffers from $\mathcal{O}(MN)$ communication cost per epoch, while \textbf{RFL-ADMM-V} reduces it to $\mathcal{O}(\sum_{i=1}^M n_i)$ (Section \ref{sec:compute-red}). 

\subsubsection{RFL} 
Each organization can have multiple clients and each client owns a horizontal table.
Below we analyze the complexity of \textbf{RFL-SGD} and \textbf{RFL-ADMM}.

\vspace{2pt}
\textbf{Computation complexity.} 
In the outer loop of the training process, the server has $\mathcal{O}(N)$ computation complexity as it computes on the logical joined table; in the inner loop, the server acts as the coordinator for each organization to perform coordinator-side SGD/ADMM computation. For \textbf{RFL-SGD}, the coordinator-side computation complexity is $\mathcal{O}(1)$, which is repeated for $N/B$ times per epoch, leading to $\mathcal{O}(MN/B)$ computation complexity for all the $M$ organizations. 
On average, the client-side computation complexity is $\mathcal{O}\left(\alpha_i / |Q_i|\right)$. 
For \textbf{RFL-ADMM}, the coordinator-side computation complexity is also $\mathcal{O}(1)$ but is repeated for $\mathcal{T}'$ times per epoch, leading to $\mathcal{O}\left(M\mathcal{T}'\right)$ for all the $M$ organizations. 
The corresponding client-side computation complexity is $\mathcal{O}\left( \mathcal{T}' n_i^q \right)$.

\textbf{Communication complexity.} 
In the outer loop of the training process, the server has the same number of communication rounds and cost as that in RFL-V;
in the inner loop, the server (coordinator) needs to gather/scatter variables from clients in each organization. 
For \textbf{RFL-SGD}, there is only one communication round for each inner loop, leading to $\mathcal{O}\left(Q\right)$ communication cost between the server and the $Q$ clients. 
Since there are $N/B$ outer loops, the total communication cost of \textbf{RFL-SGD} is $\mathcal{O}(\sum_{i=1}^M \alpha_i) + \mathcal{O}\left(QN/B\right)$. 
For \textbf{RFL-ADMM}, the number of communication rounds is $\mathcal{T}'$ for each inner loop, leading to $\mathcal{O}\left(\mathcal{T}' Q\right)$ communication cost between the server and the $Q$ clients. 
Since there is only one outer loop per epoch, its total communication cost is $\mathcal{O}\left(\sum_{i=1}^M n_i + \mathcal{T}' Q\right)$.

\subsection{Privacy Analysis} \label{sec:privacy-dp} 
To protect the privacy of the labels stored 
in the server (i.e., the dataset in Definition \ref{def:dp} refers to the label set), we leverage the existing Laplace DP mechanism~\cite{dwork2014algorithmic} that adds one-time Laplace noise with standard deviation $\lambda$ to each coordinate of the labels before training~\cite{malek2021antipodes}, so that the labels satisfy $\epsilon$-label DP (with $\delta=0$).
\begin{restatable}{theorem}{thmdp}
\label{theo:labeldp_guarantee} (Privacy guarantee for labels.)
Following standard Laplace mechanism (\cite{dwork2014algorithmic}),
$\epsilon$-label DP can be achieved by injecting additive Laplace noise with per-coordinate standard deviation $\lambda=\frac{2\sqrt{2}}{\epsilon}$.
\end{restatable}

To formally protect the privacy of client's local data, 
when updating the local model of each client, we clip the per-sample gradient with $\ell_2$-norm threshold $C$ and add Gaussian noise sampled from $\mathcal{N} (0, C^2\sigma^2)$. We accumulate privacy budget based on moments accountant in~\cite{abadi2016deep} along with training as follows:
\begin{restatable}{theorem}{thmdp} 
\label{theo:dp_guarantee}(Privacy guarantee for features.)
There exist constants $c_1$ and $c_2$ such that, given $Q$ clients with $\tau$ local steps ($\tau=\mathcal{T}\mathcal{T}'S$) for each client, 
clipping threshold $C$, noise level $\sigma$, and batch subsampling ratio $r$, for any $\epsilon\leq c_1r^2\tau$, 
DP version of Algorithm~\ref{algo:training-process} satisfies $(\epsilon , \delta)$-DP for all $\delta \geq 0$ if we choose $\sigma \geq c_2 \frac{r \sqrt{\tau \log (1 / \delta)}}{\epsilon}$.
\end{restatable}
The proof extends the Theorem 1 in~\cite{abadi2016deep} to the DP guarantee for each client in our Algorithm~\ref{algo:training-process}, where each client performs $\tau=\mathcal{T}\mathcal{T}'S$ local DP-SGD steps.

\section{Evaluation}
We study the effectiveness and efficiency of \sys, by evaluating model accuracy and performance of SGD/ADMM atop \sys as well as counterparts in diverse scenarios. These scenarios include VFL/RFL-V/RFL, non-DP/DP, different network settings, as well as different ML models on a variety of datasets. We briefly summarize our evaluation methodology and main results as follows.

\textbf{(1) Effectiveness (i.e., model accuracy).}
We regard directly training centralized (non-federated) ML models on the joined table as the baselines, and compare SGD/ADMM atop \sys with these baselines in terms of model accuracy. The experiments show that SGD/ADMM atop \sys can achieve comparable model accuracy to the baselines in both VFL/RFL-V and RFL scenarios. 
With privacy guarantee, SGD/ADMM atop \sys suffers from up to 4.5\% lower model accuracy than the baselines due to the noises injected into data labels and model training.

\textbf{(2) Efficiency (i.e., performance).} As communication is the primary bottleneck in FL~\cite{FedAvg}, our evaluation mainly focuses on the \textit{model accuracy vs. communication time} among the SGD/ADMM algorithms atop \sys. 
Compared to SGD, ADMM takes less communication time to converge to similar model accuracy. 
Specifically, SGD/ADMM algorithms atop \sys outperform counterparts (\textbf{VFL-SGD/ADMM}) in terms of communication time.

\vspace{-0.5em}
\subsection{Experimental setup}
\begin{table}
\small
\caption{The datasets and ML models used in the evaluation, where 15\% of each dataset is used for model testing.}
  \label{dataset-table}
\vspace{-1em}
  \centering
  \begin{tabular}{lccccc}
  \toprule
   \textbf{Dataset (ML Model)} & \textbf{Table} & \textbf{\#Tuple}  & \textbf{\#Feature} & \textbf{\#Class} \\
    \midrule
    MIMIC-III (LR)   & 5 &  [35K, 2.9M]   & [2, 717] & 2  \\ 
    Yelp (BERT-Softmax)   & 3 &  [35K, 3.2M]   & [3, 775] & 5 \\
    MovieLens-1M (Linear) & 3 & [6K, 0.9M]   & [4, 52] & 5\\
    MovieLens-1M (NN)   & 3 &  [6K, 0.9M]   & [4, 52] & 5   \\ 
    \bottomrule
  \end{tabular}
  
\end{table}

\subsubsection{Datasets and models} 
We train four linear/NN (neural network) ML models on three real-world datasets for both classification and regression tasks, as summarized in Table~\ref{dataset-table}.

\textbf{(1) MIMIC-III}: This is a healthcare dataset with 46K patients admitted to ICUs at the BIDMC between 2001 and 2012~\cite{MIMIC-Data}. We leverage the scripts in MIMIC-III Benchmarks~\cite{MIMIC-paper, MIMIC-github} to extract 5 tables, including \texttt{Patients},  \texttt{Admissions}, \texttt{Stays}, \texttt{Diagnoses}, and \texttt{Events}. We perform the decompensation prediction task, which uses \emph{logistic regression} (LR) to predict whether the patient’s health will rapidly deteriorate in the next 24 hours (with 0/1 label).

\textbf{(2) Yelp}: This Yelp dataset contains 3 core tables \texttt{business}, \texttt{review}, and \texttt{user}~\cite{Yelp-dataset}. The label column is \texttt{stars} in the \texttt{review} table, denoting the user rating of 1 to 5 for businesses. To be simple, we only use the tuples of restaurants in the \texttt{business} table, and obtain 3.2M reviews and ratings from 1.2M users. We first use BERT NLP model~\cite{BERT} to extract 768-dimensional embeddings from review text and then use \emph{softmax regression} for classification.

\textbf{(3) MovieLens-1M}: This dataset contains 0.9M user ratings on about 4K movies given by 6K users~\cite{Movie-dataset}, including \texttt{movies}, \texttt{ratings}, and \texttt{users} tables.  The label column is \texttt{rating} in the \texttt{ratings} table, which denotes the user rating of 1 to 5. We use both \emph{linear regression} (Linear) and \emph{neural network} (NN) model with a hidden layer of 16 neurons to predict the review score. 
If a movie has multiple genres, there are multiple tuples for this movie in the \texttt{movies} table. 

In VFL/RFL-V, each client owns a whole table. In RFL, each table is further divided into two horizontal partitions, i.e., we have two clients in each organization.

\vspace{-0.5em}
\subsubsection{Experimental settings} \label{sec:settings} By following previous work on VFL~\cite{chen2020vafl,hu2019fdml,vepakomma2018split}, we simulate VFL/RFL-V/RFL with 1 server and $Q$ clients, on a Linux machine with 
16-core CPUs and 8 GPUs. 
The algorithms are implemented with NumPy and PyTorch.\footnote{The code is available at \href{https://github.com/JerryLead/TablePuppet}{https://github.com/JerryLead/TablePuppet.}}  
We assume that the clients are geo-distributed in US/UK, using two network settings as US-UK and US-US. 
US-UK refers to the speed of AWS instances between Oregon and London with 136ms latency and 0.42 Gbps bandwidth, while US-US denotes the speed between Oregon and Virginia with 67ms latency and 1.15 Gbps bandwidth~\cite{Latency-arxiv}. 
We run each experiment three times and report the average value.

\subsubsection{Hyperparameters} We use grid search to tune hyperparameters such as the learning rate $\eta$ for SGD and the $\rho$ for ADMM. 
For each dataset, we run 10 epochs for SGD/ADMM. 
We vary the learning rate of SGD from 0.01 to 0.5, and vary $\rho$ of ADMM from 0.1 to 2. 
We set the batch size of SGD to be 10K, as small batch size leads to too many communication rounds. 
In DP scenario, for model training with DP-SGD, we set the ($\epsilon$, $\delta$)-DP by using $\epsilon=1$, $\delta=1e$-5~\cite{abadi2016deep} with clipping threshold $C=1$. 
For label DP, we set the Laplace noise $\lambda=0.5$, which results in $\epsilon=5.6$ according to Theorem \ref{theo:labeldp_guarantee}. 
This label DP level is consistent with existing label DP works that typically designate $\epsilon$ values between 3 and 8~\cite{malek2021antipodes,ghazi2021deep}. 
We leverage PyTorch's DP tool Opacus~\cite{Opacus} to implement the DP and calculate the privacy budget.

\vspace{-0.5em}
\subsection{Effectiveness of \sys} \label{evl:effectiveness}

\begin{figure*}
\centering
\includegraphics[width=0.95\textwidth]{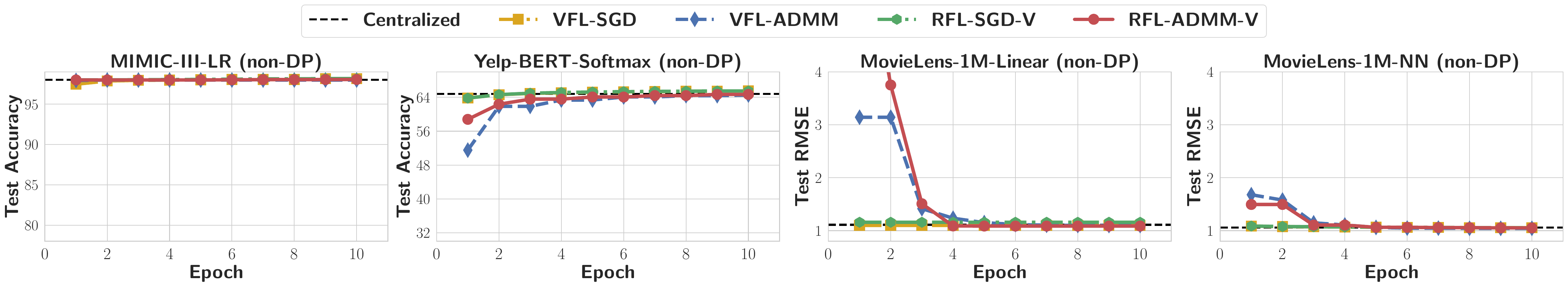}
\includegraphics[width=0.95\textwidth]{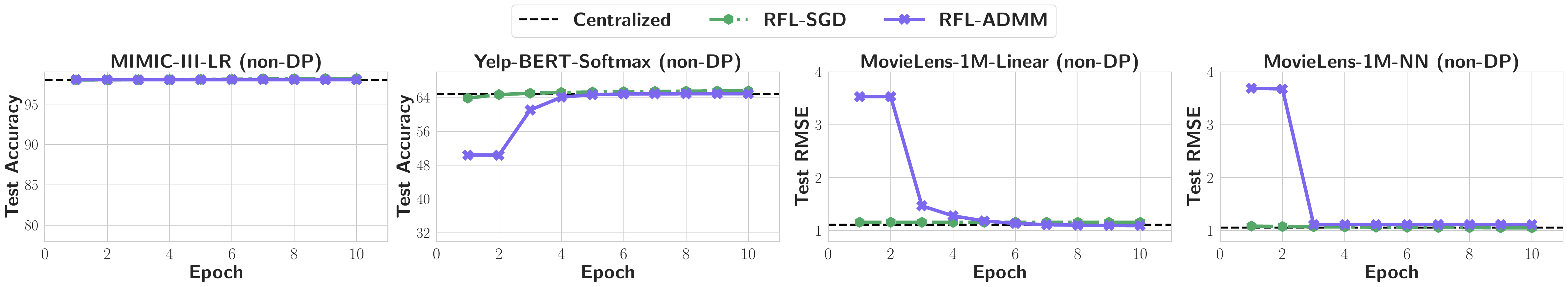}
\vspace{-1em}
\caption{The convergence rates of different algorithms for  VFL/RFL-V and RFL scenarios without privacy guarantees.}
\label{fig:convergence_rates_non-DP}

\end{figure*}

\begin{figure*}[t]
\centering
\includegraphics[width=0.95\textwidth]{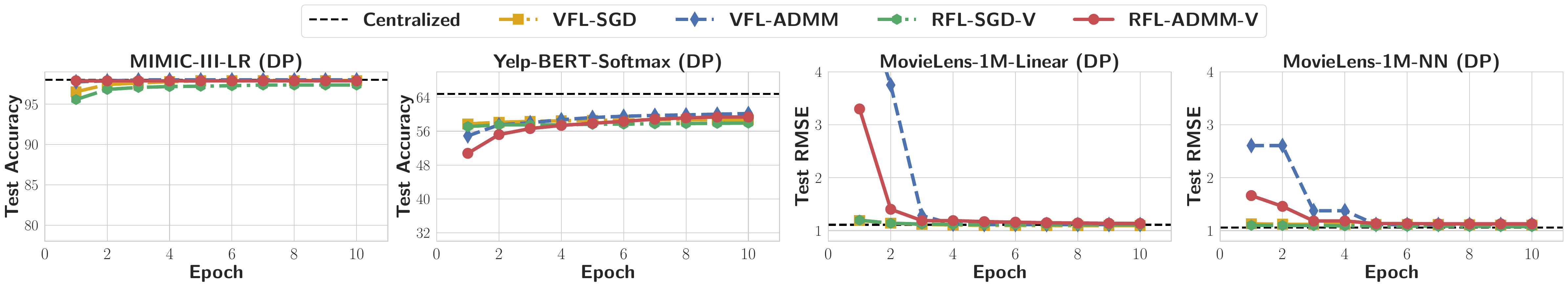}
\includegraphics[width=0.95\textwidth]{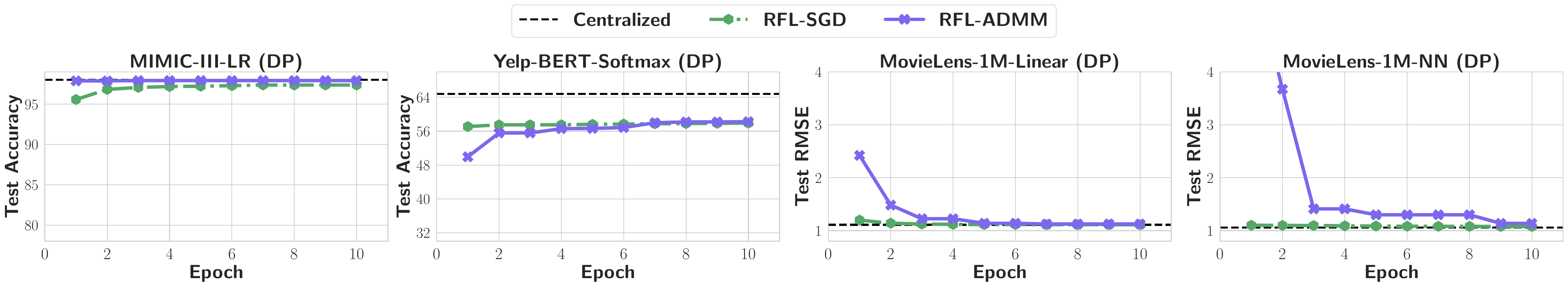}
\vspace{-1em}
\caption{The convergence rates of different algorithms for VFL/RFL-V and RFL scenarios with privacy guarantees.}
\label{fig:convergence_rates_DP}
\end{figure*}

\subsubsection{Baselines} 
As RFL is to train ML models on the UCQ result, 
we regard directly training ML models on the joined table as baselines, denoted as \textit{centralized}. 
To obtain the baseline accuracy, we use SGD to train ML models on the joined table without privacy guarantees (shown as \textit{centralized} in Figures~\ref{fig:convergence_rates_non-DP} and~\ref{fig:convergence_rates_DP}). 
The model accuracy refers to test accuracy (higher is better) for classification models and test root mean square error (RMSE, lower is better) for regression models. 
We compare the model accuracy of SGD/ADMM atop \sys with the baselines, as well as VFL methods (\textbf{VFL-SGD/ADMM}) that are forced to run directly on the vertical partitions of the
joined table.

\vspace{-0.5em}
\subsubsection{Results without privacy guarantee}
Figure~\ref{fig:convergence_rates_non-DP} illustrates the convergence rates of different SGD/ADMM algorithms without privacy guarantees. 
In this non-DP scenario, all the algorithms atop \sys can converge to model accuracy comparable to the baselines, which demonstrates the effectiveness of \sys. 

For SGD algorithms, the convergence curve of \textbf{RFL-SGD} is similar to that of \textbf{RFL-SGD-V}, because they share the same gradient computation---the computation results of Eq.~\ref{Eq:SGD-horizontal} and~\ref{Eq:sgd-client-side-horizontal1} of \textbf{RFL-SGD} are the same as Eq.~\ref{Eq:sgd-client-side-vertical1-opt} of \textbf{RFL-SGD-V}. Moreover, both of them exhibit comparable convergence rate to \textbf{VFL-SGD}, showcasing the effectiveness of our computational/communication optimization on SGD.
Compared to ADMM, SGD converges faster in most cases, because SGD updates the local models much more frequently than ADMM in each epoch. For example, on the \textit{Yelp} dataset, SGD updates local models 320 times (i.e., the number of batches) per epoch, whereas ADMM only updates the local model once per epoch (by completely solving the local optimization problem in each client). 
However, with more model updates, SGD requires more communication rounds than ADMM, leading to longer communication time per epoch (Section~\ref{evl:efficiency}). 

Regarding ADMM algorithms, \textbf{RFL-ADMM-V} achieves a similar convergence rate to \textbf{VFL-ADMM}.
This reveals that the computation/communication optimization used by \textbf{RFL-ADMM-V} is effective and does not affect the convergence of ADMM. 
Moreover, compared to \textbf{RFL-ADMM-V}, \textbf{RFL-ADMM} demonstrates a bit lower convergence rate in the first two epochs, but catches up after two epochs on \textit{Yelp} and \textit{MovieLens}.  
We speculate the reason is that \textbf{RFL-ADMM} introduces more hyper-parameters, such as the number of inner loops and the $\rho$ for horizontal ADMM, which are harder to tune than \textbf{RFL-ADMM-V}.

\vspace{-0.5em}
\subsubsection{Results with privacy guarantee}
By introducing DP to both labels and model training in \sys, the model accuracy of SGD/ADMM drops compared to the non-DP centralized baselines (Figure~\ref{fig:convergence_rates_DP}). 
Taking \textbf{RFL-SGD-V} for example, the test accuracy drops by 0.79 point for MIMIC-III-LR and 7.2 points for Yelp-BERT-Softmax, while the test RMSE slightly increases by 0.05  for MovieLens-Linear and 0.02 for MovieLens-NN. 
However, while model accuracy drops, these algorithms gain on privacy protection against feature and label leakages.
In this DP scenario, we can still observe that all algorithms atop \sys can converge to similar model accuracy. Specifically, SGD algorithms exhibit the fastest convergence in most cases due to the largest number of local model updates per epoch.

\subsection{Efficiency of \sys} \label{evl:efficiency}
We compare \textit{model accuracy vs. communication time} among SGD and ADMM algorithms. 
We presume that the server and clients are distributed in US/UK, and we use two network setups, US-UK and US-US with different latency and bandwidth, to measure the communication between the server and clients. 
Due to space constraints, we only report the US-UK results here and move the US-US results, which are similar, to our GitHub repository.
We measure the \textit{communication time} via ``\textit{latency} + \textit{communication\_data\_size} / \textit{bandwidth}'' for each epoch. 

\vspace{-0.5em}
\subsubsection{Results for VFL/RFL-V} 
\begin{figure*}
\centering
\includegraphics[width=0.95\textwidth]{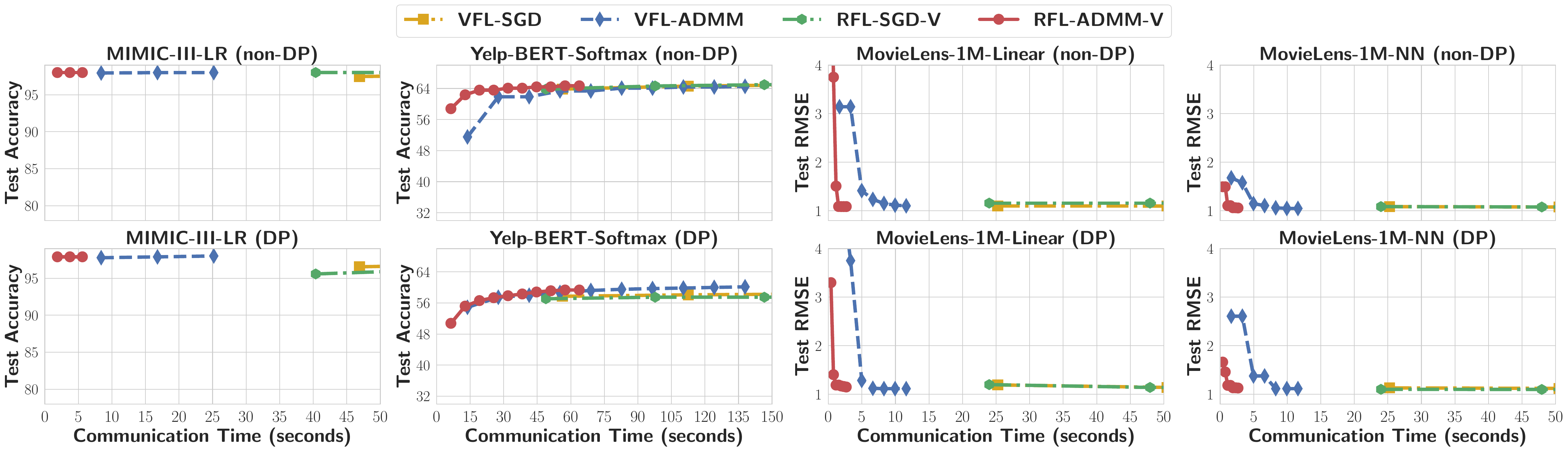}
\vspace{-0.8em}
\caption{The model accuracy vs. communication time for VFL/RFL-V (US-UK with latency = 136ms, bandwidth = 0.42Gb/s).}
\label{fig:communication_time_VFL}
\end{figure*}

Figure~\ref{fig:communication_time_VFL} illustrates the \textit{model accuracy vs. communication time} in US-UK scenario, which contains two sub-figures for non-DP (top) and DP (bottom) scenarios. 
Each point in the figure represents the test accuracy/RMSE after one epoch. 
Note that \textbf{VFL-SGD} and \textbf{RFL-SGD-V} results are not fully plotted in the figure, due to the long communication time caused by too many communication rounds per epoch.
For example, for the \textit{MIMIC-III} dataset with 2.9M tuples, \textbf{VFL-SGD} suffers from 290 communication rounds per epoch, leading to extremely long communication time ($>$130 seconds for just three epochs) that exceeds the boundary of the horizontal axis. Although \textbf{RFL-SGD-V} outperforms \textbf{VFL-SGD} in terms of communication cost, it still requires the same communication rounds that leads to the long communication time.
In contrast, \textbf{VFL-ADMM} can converge with 25s communication time, while \textbf{RFL-ADMM-V} only requires 6s. 
As another example of MovieLens-1M, \textbf{RFL-ADMM-V} converges with 4.3$\times$ less communication time than \textbf{VFL-ADMM}, owing to the communication reduction on duplicate tuples as described in Section \ref{sec:compute-red}. 
The above results are consistent with the complexity analysis in Section \ref{complexity_analysis}, where \textbf{VFL-ADMM} outperforms \textbf{VFL-SGD} due to fewer communication rounds and \textbf{RFL-ADMM-V} further outperforms \textbf{VFL-ADMM} due to less communication cost.
In addition, we observe similar results in both non-DP and DP scenarios, which indicates that the privacy guarantee does not affect the number of communication rounds as well as the communication cost.

\begin{figure*}
\centering
\includegraphics[width=0.95\textwidth]{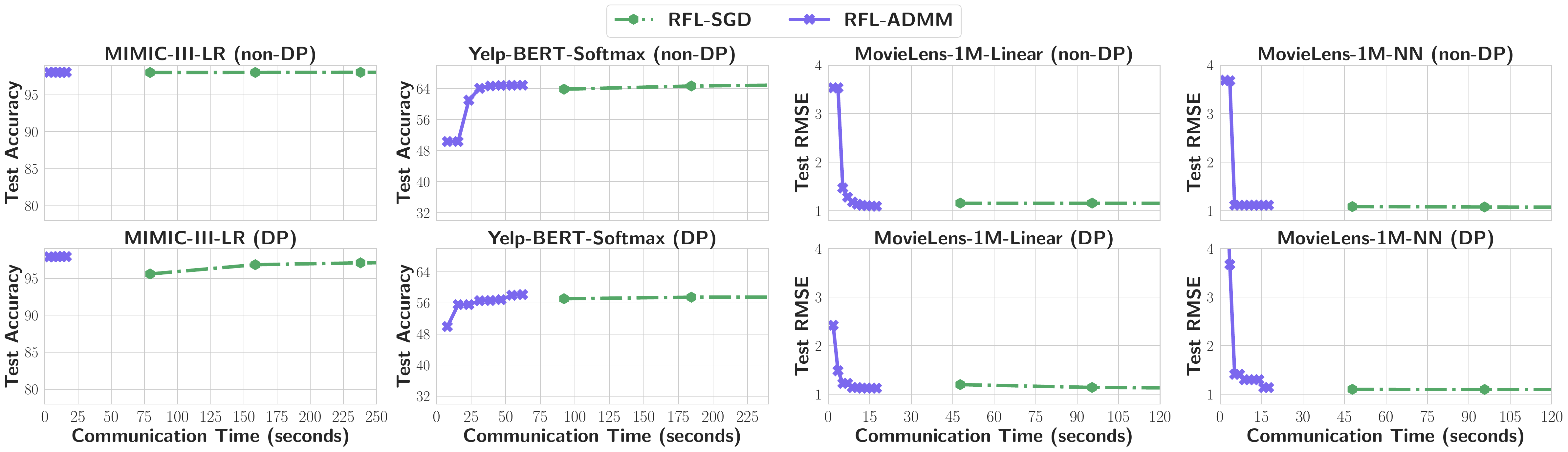}
\vspace{-0.8em}
\caption{The model accuracy vs. communication time for RFL (US-UK with latency = 136ms, bandwidth = 0.42Gb/s).}
\label{fig:communication_time_Hybrid}
\end{figure*}

\vspace{-0.5em}
\subsubsection{Results for RFL}

Figure~\ref{fig:communication_time_Hybrid} presents the \textit{model accuracy vs. communication time} in US-UK scenario. 
Although \textbf{RFL-SGD} exhibits a faster convergence rate in terms of the number of epochs (Section \ref{evl:effectiveness}), we can see that \textbf{RFL-ADMM} still takes less communication time than \textbf{RFL-SGD} to converge. 
The reason is similar to that analyzed in VFL/RFL-V, i.e., \textbf{RFL-SGD} requires more communication rounds due to per-batch communication as well as more communication costs due to duplicate tuples. 

Compared to the results in VFL/RFL-V, SGD/ADMM algorithms for RFL require longer communication time per epoch, to communicate inner-loop computation results during LoU such as partial gradients for SGD and auxiliary variables for ADMM. 
Although the size of inner-loop computation results is normally smaller than that of the outer-loop computation results for LoJ, the extra inner-loop computation increases both the communication rounds and the communication cost per epoch. 
For example, for MovieLens-1M-Linear per epoch, the extra inner-loop communication rounds/time is 175/23.35s for \textbf{RFL-SGD} and 10/1.36s for \textbf{RFL-ADMM}, respectively. 
The corresponding complexity has been analyzed in Section \ref{complexity_analysis}, where \textbf{RFL-SGD} and \textbf{RFL-ADMM} require extra $\mathcal{O}(QN/B)$ and $\mathcal{O}(\mathcal{T}' Q)$ communication cost in the inner loop. 
In summary, \textbf{RFL-ADMM-V} and \textbf{RFL-ADMM} outperform the others in terms of communication time.

\vspace{-0.5em}
\section{Related Work}

\textbf{Federated Learning (FL). } FL is a privacy-preserving technique for collaboratively training ML models across decentralized clients. Based on different types of data partitioning, existing surveys~\cite{Yang-19-survey, VFL-survey} classify FL 
into horizontal FL, vertical FL, and federated transfer learning (FTL). 
Section~\ref{sec:existing-FL} discussed horizontal and vertical FL, while FTL focuses on model transfer between clients using overlapping data samples and features~\cite{ftl, FTL-survey, chen2020fedhealth}. 
A recent addition is hybrid FL~\cite{HybridFL-arxiv,HybridFL-dual} where data can be partitioned \emph{both} horizontally and vertically; however, existing work either requires overlapping samples/features like FTL~\cite{HybridFL-arxiv} or does not support non-convex models like neural network~\cite{HybridFL-dual}. 
Moreover, current FL solutions assume that the decentralized data tables can be simply one-to-one aligned, while our RFL problem 
targets relational tables in distributed databases that require joins and unions. 
FL traditionally employs training algorithms such as SGD, ADMM, \textit{block coordinate descent} (BCD)~\cite{liu2019communication}, and \textit{gradient-boosted decision trees} (GBDT) for tree-based models~\cite{Fu-SIGMOD-Tree, wu2020privacy, FedGBF}. 
In future work, we plan to extend \sys to support more types of training algorithms in addition to SGD and ADMM.
To improve communication efficiency, we can also explore how to leverage asynchronous or other communication-efficient model update protocols from existing FL research~\cite{Fu-VLDB-cache, FederatedScope, VLDB-Hetero, async-comm}.

\vspace{3pt}
\noindent\textbf{Learning over Join. }
Several approaches have been proposed~\cite{Factorized-Joins, Layered-Aggregate, learning-GLM, linear-algebra} for learning over joins (LoJ). 
However, these approaches assume that the data tables can be joined within a single machine. 
Moreover, these approaches suffer from either generality problems (e.g., they do not support non-polynomial loss functions well, such as logistic regression model~\cite{Factorized-Joins, Layered-Aggregate, DBLP:phd-maxim}) or performance issues (e.g., they use standard \textit{gradient descent} (GD) method that is slow to converge~\cite{linear-algebra, learning-GLM, Factorized-Joins, Layered-Aggregate}). \sys supports training with the more efficient (mini-batch) SGD, which converges faster than GD. \sys also supports neural network models.

\vspace{3pt}
\noindent\textbf{Privacy guarantee for FL. } 
Different from HFL/VFL, both features and labels in RFL are distributed across clients, meaning that we need more advanced DP methods to protect them simultaneously. \sys currently offers a solution by using separate label-level DP and DP-SGD to safeguard both features and labels. It remains interesting to explore how to combine these two DP methods or incorporate other advanced label-level DP mechanisms~\cite{malek2021antipodes,ghazi2021deep}, such as post-processing the model predictions through Bayesian inference, into \sys. Moreover, cryptographic techniques like homomorphic encryption~\cite{rouhani2018deepsecure, gilad2016cryptonets} and secure multiparty computation~\cite{ben1988completeness,bonawitz2017practical} can also be potentially added into \sys to enhance its privacy guarantee, albeit with increased computation overhead.

\section{Conclusion}

We propose and formalize a new \textit{relational federated learning} problem that aims at training ML models over relational tables with joins and unions across distributed databases. 
We present \sys 
that can push two widely adopted ML training algorithms, SGD and ADMM, down to individual relational tables, with performance optimization and privacy guarantees.
The SGD/ADMM algorithms based on \sys can achieve model accuracy similar to centralized baseline approaches. Our ADMM algorithms also outperform others in terms of communication time.

\balance
\bibliographystyle{ACM-Reference-Format}
\bibliography{references, ref-vfl}

\end{document}